\documentclass[11pt,twoside]{article}
\usepackage{fullpage}

\usepackage{epsf}
\usepackage{fancyhdr}
\usepackage{graphics}
\usepackage{graphicx}
\usepackage{psfrag}
\usepackage{microtype}
\usepackage{subfigure}
\usepackage{comment}

\usepackage{enumitem}
\usepackage{algorithmic}
\usepackage{color,xcolor}
\usepackage[linesnumbered,ruled]{algorithm2e}
\DontPrintSemicolon
\usepackage{color}

\usepackage{amsthm}
\usepackage{amsfonts}
\usepackage{amsmath}
\usepackage{amssymb,bbm,bm}

\usepackage{thmtools, thm-restate}
\usepackage[colorlinks,linkcolor=magenta,citecolor=blue, pagebackref=true,backref=true]{hyperref}
\renewcommand*{\backrefalt}[4]{%
    \ifcase #1 \footnotesize{(Not cited.)}%
    \or        \footnotesize{(Cited on page~#2.)}%
    \else      \footnotesize{(Cited on pages~#2.)}%
    \fi}

\newcommand{\regret}{\mathcal{R}}

\newcommand{\real}{\ensuremath{\mathbb{R}}}

\newcommand{\statespace}{\mathcal{S}}
\newcommand{\actionspace}{\mathcal{A}}
\newcommand{\horizon}{H}
\newcommand{\transition}{P}
\newcommand{\reward}{R}
\newcommand{\MDP}{\mathcal{M}}
\newcommand{\initialdistr}{p_0}
\newcommand{\policy}{\mu}

\newcommand{\eluderdim}{\mathrm{dim}_{E}}

\newcommand{\optimalQ}{Q^*}
\newcommand{\activepolicyspace}{{\tilde{\Theta}}}

\newcommand{\thetastar}{\ensuremath{{\theta^*}}}

\newcommand{\abss}[1]{\left| #1 \right |}

\newcommand{\sphere}{\ensuremath{\mathbb{S}}}

\newcommand{\mydefn}{\ensuremath{:=}}

\newcommand{\stack}{\mathsf{S}}
\newcommand{\stackpush}{\mathsf{push}}
\newcommand{\stackpop}{\mathsf{pop}}
\newcommand{\stacktop}{\mathsf{top}}

\newcommand{\vecnorm}[2]{\left\| #1\right\|_{#2}}

\newcommand{\inprod}[2]{\ensuremath{\langle #1 , \, #2 \rangle}}

\newcommand{\Exs}{\ensuremath{{\mathbb{E}}}}
\newcommand{\Prob}{\ensuremath{{\mathbb{P}}}}

\newcommand{\Qsep}{\Delta}
\newcommand{\maxRewards}{\bar{R}}

\newtheoremstyle{named}{}{}{\itshape}{}{\bfseries}{.}{.5em}{\thmnote{#3's }#1}
\theoremstyle{named}

\theoremstyle{plain}

\newtheorem{theorem}{Theorem}
\newtheorem{proposition}{Proposition}
\newtheorem{lemma}{Lemma}

\newtheorem{definition}{Definition}

\newlength{\widebarargwidth}
\newlength{\widebarargheight}
\newlength{\widebarargdepth}

\makeatletter
\long\def\@makecaption#1#2{
        \vskip 0.8ex
        \setbox\@tempboxa\hbox{\small {\bf #1:} #2}
        \parindent 1.5em  
        \dimen0=\hsize
        \advance\dimen0 by -3em
        \ifdim \wd\@tempboxa >\dimen0
                \hbox to \hsize{
                        \parindent 0em
                        \hfil
                        \parbox{\dimen0}{\def\baselinestretch{0.96}\small
                                {\bf #1.} #2
                                }
                        \hfil}
        \else \hbox to \hsize{\hfil \box\@tempboxa \hfil}
        \fi
        }
\makeatother

\long\def\comment#1{}
\definecolor{battleshipgrey}{rgb}{0.52, 0.52, 0.51}
\definecolor{darkgray}{rgb}{0.66, 0.66, 0.66}
\definecolor{darkgreen}{rgb}{0.0, 0.2, 0.13}
\definecolor{darkspringgreen}{rgb}{0.09, 0.45, 0.27}
\definecolor{dukeblue}{rgb}{0.0, 0.0, 0.61}
\definecolor{olivedrab7}{rgb}{0.24, 0.2, 0.12}
\definecolor{darkblue}{rgb}{0.0, 0.0, 0.55}
\definecolor{darkscarlet}{rgb}{0.34, 0.01, 0.1}
\definecolor{candyapplered}{rgb}{1.0, 0.03, 0.0}
\definecolor{ao(english)}{rgb}{0.0, 0.5, 0.0}
\definecolor{applegreen}{rgb}{0.55, 0.71, 0.0}

\newtheorem{assumption}{Assumption}
\usepackage[numbers]{natbib}

\newcommand{\Qhat}{\hat{Q}}
\newcommand{\Event}{\mathcal{E}}
\newcommand{\popEvent}{\mathcal{O}}
\newcommand{\pushEvent}{\mathcal{I}}
\newcommand{\initialEvent}{\mathcal{N}}
\newcommand{\numStackEpisodes}{\sigma}
\newcommand{\Tstar}{T_*}
\newcommand{\tree}{\mathcal{T}}
\newcommand{\MDPspace}{\mathcal{H}}
\newcommand{\determTran}{F}

\newcommand{\oracle}{\mathcal{O}}

\newcommand{\featSpace}{\mathcal{C}}

\newcommand{\littlestonedim}{\mathrm{dim}_L}

\newcommand{\algorithmicexit}{\textbf{exit}}
\newcommand{\EXIT}{\STATE \algorithmicexit}

\newcommand{\gftwo}{\mathbb{F}_2}

\newtheorem{example}{Example}

\newcommand{\goodendex}{\ensuremath{\clubsuit}}

\makeatletter
\def\namedlabel#1#2{\begingroup
    #2%
    \def\@currentlabel{#2}%
    \phantomsection\label{#1}\endgroup
}
\makeatother

\definecolor{DSgray}{cmyk}{0,1,0,0}

\begin{document}

\begin{center}
{\bf{\LARGE{On the Sample Complexity of Reinforcement Learning with Policy Space Generalization}}}

\vspace*{.2in}
 {\large{
 \begin{tabular}{ccc}
  Wenlong Mou$^{\diamond}$ & Zheng Wen$^{\dagger}$ &  Xi Chen$^{\ddagger}$ 
 \end{tabular}
}}

\vspace*{.2in}

 \begin{tabular}{c}
 Department of EECS, UC Berkeley$^\diamond$
 \end{tabular}

 \vspace*{.1in}
 \begin{tabular}{c}
 Google DeepMind$^\dagger$
 \end{tabular}

 \vspace*{.1in}

 \begin{tabular}{c}
 Stern School of Business, NYU$^\ddagger$
 \end{tabular}

\vspace*{.2in}

\date{}

\vspace*{.2in}

\begin{abstract}
   We study the optimal sample complexity in large-scale Reinforcement Learning (RL) problems with policy space generalization, i.e. the agent has a prior knowledge that the optimal policy lies in a known policy space. Existing results show that without a generalization model, the sample complexity of an RL algorithm will inevitably depend on the cardinalities of state space and action space, which are intractably large in many practical problems.
    
    To avoid such undesirable dependence on the state and action space sizes, this paper proposes a new notion of eluder dimension for the policy space, which characterizes the intrinsic complexity of policy learning in an arbitrary Markov Decision Process (MDP). Using a simulator oracle, we prove a near-optimal sample complexity upper bound that only depends linearly on the eluder dimension. We further prove a similar regret bound in deterministic systems without the simulator.
\end{abstract}
\end{center}

\section{Introduction}
Recent years witness the prevailing success of reinforcement learning (RL) in various applications. The workhorses for real-world large-scale RL problems are model learning, value learning, and policy learning. For these three main categories of RL, with the required assumptions listed in a descending order, the applicability to realistic problems increases. Specifically, model learning algorithms impose assumptions and attempt to estimate the transition kernel and reward function for the entire MDP, while value learning algorithms aim at the $Q$-function; and for policy learning, only assumptions on optimal policies are needed. Policy learning algorithms make the least assumptions and avoid estimating unnecessary components of the model. The minimum-assumption nature of policy learning offers the flexibility dealing with large-scale complicated MDPs, where the model and value structures are difficult to model. In real-world applications, policy learning algorithms turn out to be effective when neural networks are used to model the policy spaces (see e.g.~\cite{levine2016end,lillicrap2015continuous,schmidhuber2019reinforcement,schulman2015trust,schulman2017proximal}, and references therein).

The key feature that enables sample-efficient learning in large-scale RL is \emph{generalization}, i.e., the learning agent generalizes past experience to state-action pairs not seen before. Without a generalization model, the sample complexity and regret will inevitably depend on the size of state-action space~\cite{azar2017minimax}, which can be prohibitively large in modern applications. A natural generalization model in the policy learning context is to assume the optimal policy lies in a known policy space. The optimal regret and sample complexity for policy learning should then depend on the ``complexity'' of the policy space itself.

From the algorithmic aspect, the de facto standard for policy learning is policy gradient methods and their relatives~\cite{kakade2002natural,konda2000actor,sutton2000policy}. However, the vanilla policy gradient method is known to suffer from the problem of insufficient exploration, namely, the policy search is done in a local and greedy way, which easily gets stuck in bad local minima. A lot of heuristics have been proposed to alleviate this issue~\cite{fortunato2017noisy,houthooft2016vime,nikolov2018information}. However, no theoretical guarantees have been provided on the performance of such exploration heuristics, and it is known (see e.g.~\cite{wen2017efficient}, Section 3) that naive randomized exploration methods such as $\varepsilon$-greedy and noise injection can be highly inefficient. The lack of sample-efficient exploration strategies may significantly limit the capability of policy learning algorithms for real-world problems.

In summary, two prominent theoretical questions exist for policy learning and generalization:
\begin{enumerate}
    \item Is there a ``intrinsic'' complexity measure of a policy space that characterizes the sample complexity of RL with policy space generalization?
    \item How to explore in an MDP with policy space generalization in a sample-efficient way?
\end{enumerate}

The main contribution of this paper is by answering both theoretical questions affirmatively. In particular, we propose a notion of eluder dimension, denoted by $\eluderdim$, for policy classes, and prove the following results:
\begin{itemize}
    \item For general MDPs where a simulator is available to start the dynamics from any starting state. We show that the sample complexity for finding an $\varepsilon$-optimal policy is upper bounded by $\tilde{O} \left( \horizon \eluderdim (\Theta) (\Delta^{-2} + \varepsilon^{-1}) \right)$, in an MDP with $\Delta$-separation for the optimal $Q$-function.
    \item For deterministic systems, we propose a learning algorithm that achieves a regret upper bound of $O (\horizon \maxRewards \cdot \eluderdim (\Theta))$ for any policy class $\Theta$ and an arbitrary finite-horizon deterministic system with horizon $\horizon$ and maximal reward $\maxRewards$.
    \item In conjunction with the upper bounds, we also prove a minimax lower bound for any given policy class, that scales linearly with the Littlestone dimension of a policy space, a weaker combinatorial notion of policy space complexity.
\end{itemize}

To provide a better understanding into aforementioned results, it is useful to relate it to existing results in this line of research. Eluder dimension was first proposed in bandit literature~\cite{russo2013eluder,russo2014learning}, which characterizes the complexity of exploration under a general framework of stochastic bandit problems, with regret upper bounds proven for UCB-type and Thompson sampling algorithms. In a general ``learning-to-optimize'' setting, the eluder dimension for a class of functions describes the longest sequence of independence, a natural measure for the number of times that an exploration algorithm can get ``eluded'' by the environment. In reinforcement learning literature, the notion of eluder dimension has been extended for model classes~\cite{osband2014model} and value function classes~\cite{wen2017efficient}.

One highly relevant literature is~\cite{wen2017efficient}, which proves upper and lower bounds for the regret in deterministic systems based on the eluder dimension of a $Q$-function class. Their results were further extended by~\cite{du2020agnostic}, which works in an agnostic setting and stochastic reward case. Compared to these prior works, our result makes the following advances: First, it generalizes the classcial eluder dimension bounds to policy learning settings, and provides an analogous theory for policy space generalization. Second, it is also important to note that under a separation assumption and a simulator, our result is valid for general MDPs. To the best of our knowledge, this is also the first sample complexity bound for policy learning, which only depends on an intrinsic complexity notion of policy spaces, without any exponential dependence on time horizon. This is also the first time that model-free notion of eluder dimension can be used to characterize the complexity of RL beyond deterministic systems. Finally, our lower bound holds true for any policy space with a given Littlestone dimension. This is in contrast to the lower bound in~\cite{wen2017efficient}, which only guarantees the existence of such function class. For more discussion with related works, see Section~\ref{app:additional-related-works}.

\subsection{Preliminaries and Problem Setup}

  We consider finite-horizon MDPs in both deterministic and stochastic settings. An MDP is defined by the tuple $(\statespace, \actionspace, \horizon, \transition, \reward, \initialdistr)$, where $\statespace$ is the state space, $\actionspace$ is the action space, and $\horizon$ is the total horizon length for the system. In stochastic systems, the transition kernel $\transition (s, a, h)$ is a probability distribution over $\statespace$ for any $s \in \statespace, a \in \actionspace, h \in [\horizon]$; the reward function $\reward (s, a, h)$ is a non-negative and bounded random variable for each $s \in \statespace, a \in \actionspace, h \in [\horizon]$; and the initial distribution $\initialdistr$ is a probability distribution over $\statespace$. In deterministic systems, $\transition$ is a function from $\statespace \times \actionspace \times [\horizon]$ to $\statespace$, $\reward$ is a deterministic function on $\statespace \times \actionspace \times [\horizon]$, and $\initialdistr$ is an atomic distribution supported on $s_0$.
  
    It is well-known that under mild conditions, there always exists a deterministic optimal policy. So we restrict our attention to deterministic policies. A policy $\policy$ is defined as a deterministic mapping from $\statespace \times [\horizon]$ to $\actionspace$, which specifies the choice of action at any state and any time. Denote by $\policy_*$ the optimal policy for the MDP. We define $\maxRewards \mydefn \sup_{\policy} \left( \mathrm{ess} \sup \sum_{h = 1}^{\horizon} \reward (s_h^\policy, a_h^\policy, h) \right) $.
  For any possible policy, the total reward is almost surely bounded by $\maxRewards$. We consider parametrized policies $(\policy_\theta)_{\theta \in \Theta}$, a class of policies indexed by $\theta$. We denote by $\policy_\Theta(s, h)$ the set $\{\policy_\theta (s, h): \theta \in \Theta\} \subseteq \actionspace$.
  
  Throughout the paper, We use $\optimalQ: \statespace \times \actionspace \times [\horizon] \rightarrow \real$ to denote the optimal $Q$ function for the MDP, and use $Q^\policy: \statespace \times \actionspace \times [\horizon] \rightarrow \real$ to denote the $Q$ function for a policy $\policy$. Similarly, we use $V^*$ and $V^\policy$ to denote the optimal value function and the value function for policy $\policy$.
  
    For $T$ episodes of the MDP $\MDP$ and an algorithm $\mathsf{Alg}$, the regret is defined as:
  \begin{align*}
      \regret_T (\mathsf{Alg}; \MDP) \mydefn \sum_{t = 1}^T \left( \Exs \optimalQ (s_0^{(t)}) - \Exs \left(\sum_{h = 0}^\horizon \reward_h^{(t)} (\mathsf{Alg}) \right)  \right),
  \end{align*}
  where $\reward_h^{(t)} (\mathsf{Alg})$ is the reward that the algorithm gets at the $h$-th epoch of episode $t$.
  
  Throughout the paper, we make the following assumption on the policy space and the MDP:
  
  \begin{assumption}\label{assume-optimal-policy-in-class}
    There exists $\thetastar \in \Theta$, such that $\policy_{\thetastar} = \policy_*$. Furthermore, the optimal policy for the MDP $\MDP$ is unique.
  \end{assumption}
  
  \subsection{Additional related works}\label{app:additional-related-works}
Exploration in model-free reinforcement learning has been intensively studied from both theoretical and practical viewpoints. In the value learning setting, a series of work~\cite{osband2017posterior,osband2017deep,russo2019worst} focuses on a class of randomized algorithms for exploration inspired by Thompson sampling. They prove Bayes and minimax regret upper bounds for the tabular setting, while experimental results show that the RLSVI algorithm works in the linear value function generalization setting. When the underlying contextual decision process satisfies a low-Bellman-rank condition, it is known~\cite{jiang2017contextual,sun2018model} that value-based and model-based algorithms can explore well with low sample complexity. Under a lower-variance condition and a distribution shift error checking oracle,~\cite{du2019provably} shows a polynomial sample complexity upper bound for learning with linear $Q$-function approximation. On the lower bound side, several exponential lower bounds are established by~\cite{du2019good} for the linear case.

Theoretical studies into policy learning algorithms have attracted attention from many different aspects. For policy gradient methods,~\cite{agarwal2019optimality} shows the convergence rate
under a low representation error condition, and a no-spurious-local-minima property is established by~\cite{bhandari2019global} under a completeness condition. Methods based on importance sampling have been studied by~\cite{papini2019optimistic,tirinzoni2019transfer}. They establish sample complexity bounds based on a variant of R\'{e}nyi divergence between the sample paths, which can be exponentially large in the planning horizon and feature dimension. Under model-based assumptions, policy learning algorithms can be analyzed with optimality guarantees. For example,~\cite{fazel2018global,malik2018derivative} establishes the convergence rate of first-order and zeroth-order policy optimization in LQRs, and~\cite{cai2019provably} shows a sample complexity bounds for policy learning algorithms under the linear MDP assumption.

\section{Combinatorial notions of policy space complexity}\label{sec:combinatorial-dimension}
In this section, we introduce and discuss the combinatorial notions of complexity used in our upper and lower bounds. We give formal definition of such notions and discuss their connections. Some illustrative examples are also provided in Section~\ref{subsec:examples}.

\begin{definition}[Distinguishability]
    Given a subset $\mathcal{Z} \subseteq \{(s, a_1, a_2, h) \in \statespace \times \actionspace^2 \times [\horizon]: a_1 \neq a_2\}$ of elements and two parametrized policies $\theta_1, \theta_2 \in \Theta$, we say that $\theta_1$ and $\theta_2$ are distinguishable with respect to $\mathcal{Z}$ if and only if:
    \begin{align*}
        \exists (s, a_1, a_2, h) \in \mathcal{Z}, \quad  a_i = \policy_{\theta_i} (s, h), ~i = 1,2 
    \end{align*}
\end{definition}
    We say $\theta_1$ and $\theta_2$ to be indistinguishable w.r.t. $\mathcal{Z}$ iff they are not distinguishable w.r.t. $\mathcal{Z}$. Intuitively, we say two policies parametrized by $\theta_1$ and $\theta_2$ are distinguishable with respect to $(s, a_1, a_2, h)$ when their $Q$-values are ``distinguished'' by this tuple, i.e., the action pair $a_1$ and $a_2$ are the actions taken by the two policies at $(s, h)$ respectively,

\begin{definition}[Dependency]
    Let $\mathcal{X} = \{(s, a_1, a_2, h): s\in \statespace, a_1, a_2 \in \actionspace, h \in [\horizon ]\}$. Given a class of policies $\{\policy_{\theta}: \theta \in \Theta \}$, $x \in \mathcal{X}$ is called dependent with $\mathcal{X}' \subseteq \mathcal{X}$ with respect to $\Theta$ if and only if the following statement holds for any pair $\theta_1, \theta_2 \in \Theta$: $\theta_1$ and $\theta_2$ are indistinguishable with respect to $\mathcal{X}'$ implies that $\theta_1$ and $\theta_2$ are indistinguishable with respect to $x$.
\end{definition}

The notion of dependency for real-valued function classes can be seen as generalization of linear dependency for vector spaces: for a $d$-dimensional linear function, the function value at a new point $x \in \real^d$ is determined by the values at $d$ linearly independent points. For binary-valued functions under our consideration, though the notion of linear dependency does not exist in general, we will see in the examples that many interesting function classes still exhibit similar structure.

Based on the notion of indistinguishablity and dependency, we define the eluder dimension:
\begin{definition}[Eluder dimension]
    Given a class $\Theta$ of policies and an MDP $\MDP$, the eluder dimension of $\Theta$ is defined as:
    \begin{align*}
        \eluderdim (\Theta) \mydefn \max \{K: ~\exists (x_i)_{1 \leq i \leq K} \in \statespace \times \actionspace^2 \times [\horizon],~ x_i \text{ is independent of $(x_j)_{1 \leq j \leq i - 1}$}\}.
    \end{align*}
\end{definition}

Finally, we define the notion of Littlestone dimension for policy classes. First introduced in~\cite{littlestone1988learning}, Littlestone dimension characterizes the online learnability of function classes. We generalize this notion to policy learning problems, and prove lower bound on the minimax regret.

In most applications, we usually associate a feature vector $\phi_{s, h} \in \featSpace$ to each $s \in \statespace, h \in [\horizon]$, where $\featSpace$ is the feature vector space. A policy $\policy_\theta$ parametrized by $\theta \in \Theta$ is a deterministic mapping from $\featSpace$ to $\actionspace$. We restrict our attention to action space $\actionspace = \{0, 1\}$ for the lower bound, in which case the complexity of the policy space can be characterized by the depth of complete binary trees.

\begin{definition}[Littlestone dimension]
    The Littlestone dimension $\littlestonedim$ of a policy class $\Theta$ is defined as the smallest $D$ such that there exists a $\featSpace$-valued complete binary tree $(\phi^{(v)})_{v \in T}$ of depth $D$, such that, for any path from root to a leaf in $T$, let the label sequence be $(b_{v_i})_{i = 1}^D$ and the feature vector sequence be $(\phi^{(v_i)})_{i = 1}^D$, there exists $\theta \in \Theta$, such that:
    \begin{align*}
        \forall i \in [D], \quad \policy_{\theta} (\phi^{(v_i)}) = b_{v_i}.
    \end{align*}
\end{definition}

It is useful to compare the two different notions of dimension in this case. Intuitively, a policy space has eluder dimension at least $D$ implies the existence of a feature sequence $(\phi^{(i)})_{1 \leq i \leq D}$ and a label sequence $(b_i)_{i = 1}^D$, such that given the decision on $(\phi^{(j)})_{1 \leq j \leq i}$, the decision on the new feature $\phi^{(i)}$ is still uncertain. Stronger conditions are needed for Littlestone dimension to be at least $D$, which require the existence of such feature vector sequence for any possible label sequence $(b_i)_{i = 1}^D$. Consequently, we have $\littlestonedim (\Theta) \leq \eluderdim (\Theta)$.

\subsection{Examples}\label{subsec:examples}

In this section, we provide illustrative examples for eluder dimension and Littlestone dimension in the policy learning context. We begin with two toy examples:
\begin{itemize}
    \item For a finite policy class $\Theta$, the eluder dimension is upper bounded by $|\Theta| - 1$.
    \item For a policy class containing all deterministic policies in the tabular setting, the eluder dimension is upper bounded by $|\statespace| \cdot |\actionspace|^2 \cdot \horizon$.
\end{itemize}

For most modern applications of policy space generalization, the policy is defined as a mapping from a feature vector $\phi \in \real^d$ associated to $(s, h) \in \statespace \times [\horizon]$ to $\actionspace$. And it is typically parametrized by a vector $\theta$. In the following three examples, we consider the case with $\actionspace = \{0, 1\}$, and the policy classes defined by linear threshold functions.

\begin{example}[Linear threshold policies with worst-case features]\label{example:linear-worst-case} \upshape
    Consider feature vector space $\real^d$, with the policy defined by:
    \begin{align*} 
        \forall \theta \in \real^d, \quad\policy_\theta (s, h) = \bm{1}_{\inprod{\theta}{\phi (s, h)} > c (s, h)}
    \end{align*}
    for feature vector $[\phi (s, h), c (s, h)]$ associated to $s \in \statespace$ and $h \in [\horizon]$. It is known~\cite{alon2019private} that linear threshold functions have infinite Littlestone dimension, and consequently also have infinite eluder dimension. See Appendix~\ref{subsec:littlestone-infinite} for more details.  \hfill \goodendex
\end{example}

\begin{example}[Linear threshold policies with $\varepsilon$-packing and random features]\upshape \label{example:linear-random}
    Consider the case of $\policy_\theta (\phi) = \bm{1}_{\inprod{\theta}{\phi} > 0}$. Though the eluder dimension is infinite for linear threshold functions in general, if the feature vectors are $\mathrm{i.i.d.}$ random Gaussian and the set $\Theta$ is a discrete approximation to a set in $\real^d$, the eluder dimension can have a polynomial upper bound, as stated in the following proposition:
    
    \begin{proposition}\label{prop:eluder-example-random-feature}
        Given $\varepsilon > 0$, let $\Theta$ be a finite subset of $\sphere^{d - 1}$, such that for $\theta_1, \theta_2 \in \Theta$, there is $\vecnorm{\theta_1 - \theta_2}{2} \geq \varepsilon$. For $(\varphi_i)_{i = 1}^{+ \infty} \sim \mathrm{i.i.d.} \mathcal{N} (0, I_d)$, with probability $1 - \delta$, the largest $D$ satisfying $\phi_i$ is independent of $(\phi_{j})_{j = 1}^{i - 1}$ for $i \in [D]$ can be upper bounded with:
        \begin{align*}
        D \leq \frac{4 \pi}{\varepsilon} \log \frac{|\Theta|}{\delta}.
        \end{align*}
    \end{proposition}
    \upshape
    See Appendix~\ref{subsec:eluder-random-feature} for the proof of the proposition.
    
    For example, if $\Theta$ is an $\varepsilon$-packing of the sphere $\sphere^{d - 1}$, the eluder dimension under $\mathrm{i.i.d.}$ Gaussian feature vectors is upper bounded with $O \left( \frac{d \log \varepsilon^{-1} + \log \delta^{-1}}{\varepsilon} \right)$ with probability $1 - \delta$. For an $\varepsilon$-packing of the set of $s$-sparse vectors in $\sphere^{d - 1}$, the bound becomes $O \left( \frac{s \log (d / \varepsilon) + \log \delta^{-1}}{\varepsilon} \right)$.
    
    If the feature vectors in the MDP are $\mathrm{i.i.d.}$ standard Gaussian, on the event that the claim in Proposition~\ref{prop:eluder-example-random-feature} holds true, the upper bounds in Theorem~\ref{thm:stochastic-main} and~\ref{thm-deterministic-main} are also valid.\hfill \goodendex
\end{example}

Comparing Example~\ref{example:linear-worst-case} and Example~\ref{example:linear-random}, it suggests taking an $\varepsilon$-net, instead of the entire parameter space, can be helpful for generalization in policy spaces, with the help of random feature vectors.

In many applications with discrete input features, it is natural to model the problem in a finite field instead of $\real^d$. As discussed in the following example, for the family of linear functions in a Galois field, the eluder dimension and Littlestone dimension are both exactly the dimension of the vector space.

\begin{example}[Linear functions in Galois field] \label{example:gf2}  \upshape
    Let $\gftwo = \{0, 1\}$ denote the Galois field with two elements. Consider the feature space $\featSpace = \gftwo^D$ for some $D > 0$, and let the parameter space be $\Theta = \gftwo^D$. Let the policy be defined as:
    \begin{align*}
        \forall \theta \in \gftwo^D, \quad \policy_\theta (\phi) = \inprod{\theta}{\phi} = \sum_{i = 1}^D \theta_i \phi_i,
    \end{align*}
    where the addition and multiplication are defined under $\gftwo$. The action space $\actionspace = \{0, 1\}$ here should also be interpreted as elements in $\gftwo$. Apparently, the policy class $\Theta$ has cardinality $2^D$. The eluder dimension, however, can be much smaller, as stated in the following proposition:
    \begin{proposition}\label{prop:eluder-gf2}
        For the class of $D$-dimensional linear functions in $\gftwo$, we have:
        \begin{align*}
            \eluderdim (\Theta) = \littlestonedim (\Theta) = D.
        \end{align*}
    \end{proposition}
    See Appendix~\ref{subsec:proof-gf2} for a proof of this claim. \hfill \goodendex
\end{example}
 As we will see in two next sections, in this example, not only the bounds in Theorem~\ref{thm:stochastic-main} and Theorem~\ref{thm-deterministic-main} significantly reduces the sample complextiy, but they are also optimal up to $\horizon$ factors, according to Theorem~\ref{thm-deterministic-lower-bound}.

 \begin{example}[Boolean functions with small fourier support]\label{example:discrete-fourier} \upshape
     We consider feature space $\featSpace = \{-1, 1\}^D$ and Boolean functions $f: \featSpace \rightarrow \{-1, 1\}$. It is known~\cite{o2014analysis} that a Boolean function can be represented by its Fourier coefficients:
     \begin{align*}
         f (x) = \sum_{S \subseteq [D]} \chi_S (x) \hat{f} (S),
     \end{align*}
     where $\chi_S (x) \mydefn \prod_{i \in S} x_i$ are the basis functions and $\hat{f} (S) \mydefn \Exs_{X \sim \mathrm{Unif}} [f (X) \chi_S (X)]$.
     
     We consider the set of Boolean functions whose Fourier coefficients are supported on a subset $\mathcal{A}$ of $2^{[D]}$. In particular, we define the function class:
     \begin{align}
         \mathcal{F} (\mathcal{A}) \mydefn \left\{f : \featSpace \rightarrow \{-1, 1\}, ~\mathrm{s.t.} \hat{f} (S) = 0 ,~ \forall S \subseteq \mathcal{A} \right\}. \label{eq:defn-fourier-concentration}
     \end{align}
     The Fourier coefficients of many important functions are concentrated on low-degree subsets. For example, a decision tree of depth $k$ has Fourier coefficients supported on degree at most $k$. The following proposition provides an upper bound on the eluder dimension of $\mathcal{F} (\mathcal{A})$:
     \begin{proposition}\label{prop:eluder-fourier}
        For any $\mathcal{A} \subseteq 2^{[D]}$ and the function class $\mathcal{F} (\mathcal{A})$ defined in Eq~\eqref{eq:defn-fourier-concentration}, we have:
        \begin{align*}
            \eluderdim (\mathcal{F} (\mathcal{A})) \leq |\mathcal{A}|.
        \end{align*}
         In particular, when $\mathcal{A}$ contains sets of size at most $k$, we have $|\mathcal{A}| \leq \binom{D}{0} + \binom{D}{1} + \cdots + \binom{D}{k} \leq (e D / k)^k$. 
     \end{proposition}
     See Appendix~\ref{subsec:proof-fourier-boolean} for the proof of this claim.
     
     Note that the number of functions in $\mathcal{F} (\mathcal{A})$ can grow exponentially with $|\mathcal{A}|$ (for example, when $\mathcal{A} = 2^{[D]}$, we have $|\mathcal{F} (\mathcal{A})| = 2^{2^D}$), while the eluder dimension bound is linear in $|\mathcal{A}|$. Furthermore, if we use Proposition~\ref{prop:eluder-fourier} as a coarse upper bound for the eluder dimension of depth-$k$ decision trees, this leads to a non-trivial upper bound of order $O (D^k)$ for $k \ll D$. Note that this is much smaller than the number of such decision trees, which scales as $O (D^{2^k})$.
     \hfill \goodendex
 \end{example}

\section{Policy Space Generalization in Stochastic Systems}

In this section, we present our main results for policy space generalization in general MDPs. Throughout this section, we assume an access to a \emph{simulator}, i.e., the learning agent can choose to start the MDP at any state and any time horizon. We first introduce the policy learning algorithm, and then show an upper bound for its sample complexity guarantee.

\subsection{A policy learning algorithm for finite-horizon MDPs}

In this section, we present the algorithm that interacts with the MDP environment using a simulator. Before describing the algorithms and regret bounds, it is useful to clarify the assumptions and certain quantities used in the algorithm.

First, we make the following assumption on the structure of the policy space:
\begin{assumption}\label{assume-Q-separation}
    For any $s \in \statespace$ and $h \in [\horizon]$, there is:
    \begin{align*}
        \optimalQ (s, \policy_* (s, h), h) \geq \max_{a' \in \policy_\Theta (s, h) \setminus \{ \policy_* (s, h) \}}  \optimalQ (s, a', h) + \Qsep.
    \end{align*}
\end{assumption}

Assumption~\ref{assume-Q-separation} asserts a positive separation on the optimal $Q$-function between the action taken by the optimal policy and any other action that could be taken by the policies in the policy space. Intuitively, the gap makes it possible to eliminate undesirable actions with high confidence and generalize to other states. The $Q$-function gap assumption also is used in many existing analyses of RL algorithms, including tabular setting~\cite{tewari2008optimistic,simchowitz2019non} and function approximation~\cite{du2019provably,du2020agnostic}. Since our Assumption~\ref{assume-Q-separation} is defined with respect to the actions that could be chosen by the policy space, instead of the entire action space of the MDP, this assumption could be weaker than existing works.

To simplify the presentation, we make a slight modification on the MDP $\MDP$: we assume that the initial state $s_0$ is fixed, with only one available action $a_0 \in \actionspace$. We note that this assumption can be made without loss of generality: in particular, given an MDP with initial distribution $\initialdistr$, we can add an initial state $s_0$ at the beginning of the MDP, and the transition kernel under $(s_0, a_0, 0)$ is defined as $p_0$. This modified MDP is equivalent to the original MDP, with horizon larger by $1$.

Now we are ready to introduce the algorithms for policy learning. Algorithm~\ref{alg:policy-elimination-simulator} is the main algorithm that interacts with the environment. It uses an exploration procedure specified by Algorithm~\ref{alg:stochastic-subroutine}.

\begin{algorithm}[htb]
    \caption{Policy Elimination in general MDPs with a Simulator}\label{alg:policy-elimination-simulator}
    \begin{algorithmic}[1]
    \REQUIRE Oracle access to the MDP $\MDP$ starting from any $(s, a, h) \in \statespace \times \actionspace \times [0, \horizon]$, number of iterations $\Tstar$, and number of inner loop sample paths $N$.
    \ENSURE An $\varepsilon$-optimal policy $\hat{\policy}$.
    \STATE Initialize with $\Qhat (s, a, h) = 0$ for any $s \in \statespace, a \in \actionspace, h \in [0, \horizon]$ and $\activepolicyspace_1 = \Theta$, stack $\stack = \varnothing$.
    \FOR{$k = 1, 2, \cdots, \Tstar$}
        \IF{$\stack \neq \varnothing$}
        \STATE let $(s, a_1, a_2, \tilde{h}) = \stack.\stacktop$.
        \STATE Run Algorithm~\ref{alg:stochastic-subroutine} with starting state $(s, \tilde{h})$ and initial action $a_1$, stack $\stack$. \label{algstep:first-call-subroutine}
        \IF{$\stack.\stacktop$ was not changed}
            \STATE Run Algorithm~\ref{alg:stochastic-subroutine} with starting state $(s, \tilde{h})$ and initial action $a_2$, stack $\stack$.\label{algstep:second-call-subroutine}
        \ENDIF
    \IF{$\stack.\stacktop$ was not changed} 
        \STATE Perform $\stack.\stackpop$. \label{algstep:stochastic-stackpop}
        \STATE Let $j = \arg\min_{j' \in \{1, 2\}} \Qhat (s, a_{j'}, h)$.
        \STATE Update $\activepolicyspace_{k + 1} = \activepolicyspace_k \cap \{\theta: \policy_\theta (s, h) \neq a_j\}$.\label{algstep:stochastic-elimination}
    \ELSE 
    \STATE Keep the active policy space unchanged: $\activepolicyspace_{k + 1} = \activepolicyspace_k$.
      \ENDIF
      \ELSE
        \STATE Run Algorithm~\ref{alg:stochastic-subroutine} from $(s_0, a_0, 0)$ with stack $\stack$.
      \ENDIF
    \ENDFOR
    \RETURN Any policy $\theta \in \activepolicyspace_{\Tstar}$.
    \end{algorithmic}
\end{algorithm}
Algorithm~\ref{alg:policy-elimination-simulator} is based on elimination in the policy space, by keeping track of a set $\activepolicyspace_k$ of active policies at round $k$. When an algorithm is confident about the superiority of one action over another at $(s, h) \in \statespace \times [\horizon]$, a reduction of the policy space can be carried out. However, there are many possible locations about which the current active policy space is uncertain, and the decisions at those locations can have inter-dependence. All of them serve as potential candidates for elimination, yet the important thing is to choose the order of elimination. As an informal heuristics, the agent should solve the ``easier'' problems first, whose solution do not depend on the decision at another state.

\begin{algorithm}[htb]
    \caption{Exploration in Stochastic Systems} \label{alg:stochastic-subroutine}
    \begin{algorithmic}[1]
    \REQUIRE Starting point $(s_{h_0}, h_0)$, stack $\stack$, active policy space $\activepolicyspace$, initial action $a_{h_0}$. Number of sample paths $N$.
    \FOR{$i = 1, 2, \cdots, N$.}
    \STATE Take action $a_{h_0}$ and observe $(s_{h_0 + 1}^{(i)}, \reward_{h_0 + 1}^{(i)})$.
      \FOR{$h = h_0 + 1, h_0 + 2, \cdots, \horizon$}
      \STATE Choose $a_h^{(i)} = \arg\max_{a \in \policy_{\activepolicyspace_k} (s_h^{(i)}, h)} \Qhat (s_k^{(i)}, a, h)$.\label{algstep:visit-active-space-1}
      \STATE Observe state transition $s_{h + 1}^{(i)}$ (if $h < \horizon$) and reward $\reward_h^{(i)}$.
      \ENDFOR
       \IF{$\exists h \in [h_0, H]: |\policy_{\activepolicyspace_k} (s_{h}^{(i)}, h)| > 1$} \label{algstep:visit-active-space-2}
      \STATE Let $\tilde{h}$ be the largest such $h$, $a_1 \mydefn a_{\tilde{h}}^{(i)}$, and choose $a_2 \in \policy_{\activepolicyspace_k} (s_{\tilde{h}}^{(i)}, \tilde{h}) \setminus \{a_1\}$. \label{algstep:visit-active-space-3}
      \STATE $\stack.\stackpush(s_{\tilde{h}}^{(i)}, a_1, a_2, \tilde{h})$. \label{algstep:stochastic-stackpush}
      \EXIT
      \ENDIF
    \ENDFOR
      \STATE Update the $Q$ function estimator $\Qhat (s_{h_0}, a_{h_0}, h_0) \mydefn \frac{1}{N} \sum_{i = 1}^{N} \left( \sum_{h = h_0}^{\horizon - 1} \reward_h^{(i)} \right)$.\label{algstep-stochastic-q-update}
    \end{algorithmic}
\end{algorithm}

To put the heuristics into rigorous proofs, Algorithm~\ref{alg:policy-elimination-simulator} uses a \emph{stack} to maintain the information about states and actions where the exploration is needed for the current active policy space. An element in the stack is a tuple of $(s, a_1, a_2, h) \in \statespace \times \actionspace^2 \times [\horizon]$, which stands for the uncertainty of current active policy space at $(s, h)$ between actions $a_1$ and $a_2$. Until further changes being made in the stack, the exploration procedures at this state. When a new uncertain tuple is discovered at a later horizon, the new element is pushed into $\stack$. When the policy elimination operation is performed on the top element, it is popped from the stack. The FILO nature of the stack automatically guarantees that the elimination can be done only when no further uncertainty exists in the path with high probability, and the optimal policy will always remain safe.

Now we take a closer look at Algorithm~\ref{alg:stochastic-subroutine}: it explores the MDP based on the current active policy space, and when some uncertainty is encountered, it pushes the last uncertain element into the stack and exits. This procedure is repeated for $N$ times in order to be confident about the non-existence of further uncertainty. (Actually, an uncertain state later in the path might still exists, but is met with small probability). Finally, if no uncertain elements are met in the path, the algorithm is able to estimate the $Q$-function accurately, which serves as the criteria for policy elimination.

\subsection{Efficient Implementation}
In the description of Algorithm~\ref{alg:policy-elimination-simulator} and~\ref{alg:stochastic-subroutine}, we explicitly use the active policy spaces $\activepolicyspace_k$ at each iteration. In practice, the policy space can be very large or even infinite, making the na\"{i}ve implementation computationally inefficient. In this section, we describe a computationally efficient implementation of the policy elimination algorithms, without keeping track of the entire policy space.

For efficient implementation, we require the following oracle for the underlying policy space $\Theta$:
\begin{definition}[Elimination oracle]\label{def:elimination-oracle}
    For a policy space $\Theta$, an elimination oracle $\oracle$ takes a finite set $Z = \{(s_k, a_k, h_k)\}_{k = 1}^M \subseteq \statespace \times \actionspace \times [\horizon]$ and a pair $(s', h') \in \statespace \times [\horizon]$ as input, and output the following subset of $\actionspace$:
    \begin{align*}
        \oracle (Z; (s', h')) \mydefn \{\policy_\theta (s', h'): \theta \in \Theta, ~\mathrm{s.t.}~ \forall k \in [M], \policy_{\theta} (s_k, h_k) \neq a_k\} 
    \end{align*}
\end{definition}
We note that this oracle can be implemented using $O (|\statespace| \cdot |\actionspace| \cdot H \cdot M)$ time and space in the tabular case, by simply checking whether $(s', h')$ equals one of $(s_k, h_k)_{k = 1}^M$ and eliminating the corresponding actions. In the examples of $d$-dimensional linear threshold function and linear function over $\gftwo$, the elimination oracle can also be efficiently implemented in $\mathrm{poly} (d, M)$ time and space. See Appendix~\ref{app:efficient-oracle-examples} for details.

Given an elimination oracle, the algorithms can be implemented efficiently, as stated in the following proposition:
\begin{proposition}\label{prop:efficient-implementation}
    Given $(\Tstar, N)$, Algorithm~\ref{alg:policy-elimination-simulator} can be implemented by $O (\Tstar N H)$ calls to the elimination oracle, each with $M \leq \Tstar$, and additional time and space polynomial in $(\Tstar, N, H)$.
\end{proposition}
The proof of this proposition is straightforward: throughout the execution of the algorithm, a set $Z_k$ is kept, such that:
\begin{itemize}
    \item Initialization with $Z_1 = \varnothing$.
    \item When the active policy space  line~\ref{algstep:stochastic-elimination} in Algorithm~\ref{alg:policy-elimination-simulator} is executed, we update with $Z_{k + 1} = Z_k \cup \{(s, a_j, h)\}$ (otherwise, $Z_{k + 1} = Z_k$, which is unchanged).
    \item When the active policy space $\activepolicyspace_k$ is used (line~\ref{algstep:visit-active-space-1},~\ref{algstep:visit-active-space-2},~\ref{algstep:visit-active-space-3} in Algorithm~\ref{alg:stochastic-subroutine}), call the elimination oracle on the current set $Z_k$ and the state-time pair $(s_h^{(i)}, h)$ encountered. Use the output of elimination oracle in the place of the desired action set $\policy_{\activepolicyspace_k} (s_h^{(i)}, h)$.
\end{itemize}
It can be easily seen that $\activepolicyspace_k = \{\theta \in \Theta: \forall (s,a,h) \in Z_k, \policy_\theta (s, h) \neq a\}$ for each $k \geq 0$, and therefore the above procedure is valid. To study the time and space complexity, we note that the elimination step~\ref{algstep:stochastic-elimination} can be executed at most once in each iteration of Algorithm~\ref{alg:policy-elimination-simulator}, and therefore $|Z_k| \leq \Tstar$ for any $k \leq \Tstar$. On the other hand, note that the $(s, h)$ pair visited by the algorithm is upper bounded with $O (\Tstar N H)$. The number of oracle calls can be thus upper bounded as well. Finally, to implement the algorithm, one needs to store the set $Z_k$ and the $Q$ function values at all the $(s, a, h)$ tuples visited, which is also upper bounded by $O (\Tstar N H)$.

\subsection{Sample complexity guarantees}
In Theorem~\ref{thm:stochastic-main}, we present the sample complexity bounds for Algorithm~\ref{alg:policy-elimination-simulator}. Since the MDP is assumed to start from a fixed state $s_0$, as discussed in the previous subsection, we only need to look at the value function at $s_0$.

\begin{theorem}\label{thm:stochastic-main}
    Given a policy space $\Theta$ and $\varepsilon, \delta > 0$, under Assumption~\ref{assume-Q-separation} and Assumption~\ref{assume-optimal-policy-in-class}, let $\Tstar \mydefn 6 \horizon + 4 (\horizon + 1) \eluderdim (\Theta)$ and $N \mydefn \frac{8 }{\Qsep^2} \maxRewards^2 \log (\frac{4 \Tstar}{\delta}) + \frac{8 \maxRewards }{\varepsilon}$. With probability $1 - \delta$, running $\Tstar$-rounds of Algorithm~\ref{alg:policy-elimination-simulator} with parameter $(\Tstar, N)$ output an active policy space $\activepolicyspace_{\Tstar}$ satisfying:
    \begin{align*}
        \forall \theta \in \activepolicyspace_{\Tstar}, \quad V^{\policy_\theta} (s_0, 0) \geq V^* (s_0, 0) - \varepsilon.
    \end{align*}
\end{theorem}
See Section~\ref{app:stochastic-proof} for the proof of this theorem.

If we count the total number of sampled trajectories in the MDP, the sample complexity of Algorithm~\ref{alg:policy-elimination-simulator} is $O \left( \horizon D \left( \frac{1}{\Qsep^2} \log \frac{\horizon D}{\delta} + \frac{1}{\varepsilon} \right) \right)$, for a policy class with eluder dimension $D$. The sample complexity does not depend on the size of state-action space, but depends only on the intrinsic complexity $\eluderdim (\Theta)$, the time horizon, the separation in optimal $Q$ value, and the desired accuracy level. The $O (\Qsep^{-2})$ dependency is not improvable in general, as in the case of classical best-arm-identification problems. It is worth noticing that the sample complexity has a linear dependence on $\varepsilon^{-1}$. This is because by observing $N$ sample paths without a bad event, we can guarantee the probability of the bad event to be upper bounded by $O( 1/ N)$. The bound is in accordance with the fast rates for classification problems in the realizable setting. As a caveat, we note that the parameter choice in Theorem~\ref{thm:stochastic-main} depends on $\Qsep$, $\varepsilon$, and $\eluderdim (\Theta)$, which may not be known in practice. An important future direction is to make the algorithm adaptive to those parameters.

\section{Low-regret Policy Learning in Deterministic Systems}

Note that Algorithm~\ref{alg:policy-elimination-simulator} requires a simulator in order to visit a state that has been visited before, making it possible to compare two actions at the same state. When the underlying system is deterministic, we can simply record the path that leads to a state at an epoch, and take the same path to visit this state again. The necessity of a simulator can be avoided. Furthermore, without the estimation error, the algorithm can eliminate policies \emph{without the separation condition in $\optimalQ$}. So Assumption~\ref{assume-Q-separation} is \emph{not} needed either. In this section, we describe a learning algorithm in deterministic systems, and show its regret upper bound based on the eluder dimension. We also present a minimax lower bound for a given policy class, based on its Littlestone dimension.

\subsection{Policy learning algorithms in deterministic systems}

The policy learning algorithm that interacts with the deterministic system is described in Algorithm~\ref{alg:deterministic}. It uses an exploration subroutine, which is shown in Algorithm~\ref{alg:optimistic-subroutine}. The learning algorithm for deterministic systems overall resembles Algorithm~\ref{alg:policy-elimination-simulator} and Algorithm~\ref{alg:stochastic-subroutine}, albeit the simulator is not needed.

\begin{algorithm}[htb]
    \caption{Deterministic Policy Elimination}\label{alg:deterministic}
    \begin{algorithmic}[1]
    \STATE Initialize with $\Qhat (s, a ,h) = 0$ for any $s \in \statespace, a \in \actionspace, h \in [\horizon]$ and $\activepolicyspace_1 = \Theta$.
    \STATE Initial stack $\stack = \varnothing$.
    \FOR{$k = 1, 2, \cdots$}
    \IF{$\stack = \varnothing$}
        \STATE Run Algorithm~\ref{alg:optimistic-subroutine} from $(s_0, 1)$ with stack $\stack$.
    \ELSE
    \STATE Let $(s, a_1, a_2, \tilde{h}) = \stack.\stacktop$.
    \STATE Take the same sequence of actions as previous episodes to reach state $s$ at time $\tilde{h}$, and generate the sequence $(a_h^{(k)}, s_{h + 1}^{(k)}, \reward_h^{(k)})_{h = 1}^{\tilde{h} - 1}$.
    \STATE Take action $a_{\tilde{h} + 1}^{(k)} = a_2$, and receive state $s_{\tilde{h} + 1}^{(k)}$.
    \STATE Run Algorithm~\ref{alg:optimistic-subroutine} starting from $(s_{\tilde{h} + 1}^{(k)}, \tilde{h} + 1)$ with stack $\stack$.
    \IF{$\stack.\stacktop$ does not change}
        \STATE Perform $\stack.\stackpop$.
        \STATE Let $j = \arg\max_{j' \in \{1, 2\}} \Qhat (s, a_{j'}, h)$ and $i = \{1, 2\} \setminus \{j\}$.
        \STATE Update $\activepolicyspace_{k + 1} \mydefn \activepolicyspace_k \cap \{\theta: \policy_\theta (s, \tilde{h}) \neq a_i\}$. \label{algstep:deterministic-policy-space-elimination}
        \IF{$\exists a' \neq a_j$ such that $a' \in \policy_{\activepolicyspace_{k + 1}} (s, \tilde{h})$}
        \STATE $\stack.\stackpush (s, a_j, a', \tilde{h})$.\label{algstep:deterministic-push-in-tournament}
        \ENDIF
    \ENDIF
      \ENDIF
    \ENDFOR
      \end{algorithmic}
\end{algorithm}

\begin{algorithm}[htb]
    \caption{Exploration in Deterministic Systems}
    \label{alg:optimistic-subroutine}
    \begin{algorithmic}[1]
    \REQUIRE Starting point $(s_{h_0}^{(k)}, h_0)$, stack $\stack$, active policy space $\activepolicyspace_k$
    \FOR{$h = h_0, h_0 + 1, \cdots, \horizon $}
      \STATE Choose $a_h^{(k)} = \arg\max_{a \in \policy_{\activepolicyspace_k} (s_h^{(k)})} \Qhat (s_k^{(h)}, a, h)$.
      \STATE Observe state transition $s_{h + 1}^{(k)}$ and reward $\reward_h^{(k)}$.
      \ENDFOR
      \IF{$\exists h \in [h_0, H - 1],: |\policy_{\activepolicyspace_k} (s_{h}^{(k)}, h)| > 1$}
      \STATE Let $\tilde{h}$ be the largest such $h$ and choose any $a_2 \in \policy_{\activepolicyspace_k} (s_{h}^{(k)}, h) \setminus \{a_{\tilde{h}}^{(k)}\}$
      \STATE $\stack.\stackpush(s_{\tilde{h}}^{(k)}, a_{\tilde{h}}^{(k)}, a_2, \tilde{h})$. \label{algstep:deterministic-subroutine-push-when-new}
      \ENDIF
      \STATE Update $\Qhat (s_{\horizon}^{(k)}, a_{\horizon}^{(k)}, \horizon)  = \reward_{\horizon}^{(k)}$.
        \FOR{$h = \horizon - 1, \cdots 1, 0$}
       \STATE Update the $Q$ function estimator $\Qhat (s_h^{(k)}, a_h^{(k)}, h) = \reward_h^{(k)} + \max_{a' \in \policy_{\activepolicyspace_{k}} (s_{h + 1}^{(k)})} \Qhat (s_{h + 1}^{(k)}, a', h + 1)$ \label{algstep:deterministic-q-updates}
      \ENDFOR
      \end{algorithmic}
\end{algorithm}

We first note that the algorithms do not require any simulator. In particular, since the starting state and the transitions are fixed in deterministic systems, if we want to reach a state that has been seen before, we can just take the sequence of actions taken before that leads to the state. Additionally, compared to Algorithm~\ref{alg:policy-elimination-simulator} and Algorithm~\ref{alg:stochastic-subroutine}, there are two major differences: first, the exploration procedure in Algorithm~\ref{alg:optimistic-subroutine} does not require multiple trials starting from a state. This is because no estimation error is incurred by the $Q$ function estimator. Second, for any item $(s, a_1, a_2, h)$, we always maintain the fact that the path from $(s, h)$ through action $a_1$ has no uncertainty afterwards. This is because we choose the latest epoch where no uncertainty exists. When we inspect the top element of the stack, there is no need to try the path from $a_1$ again.

\subsection{Regret guarantees}

The regret guarantee for the algorithm is presented in the following theorem.

\begin{theorem}\label{thm-deterministic-main}
    Given a policy class $\Theta$, under Assumption~\ref{assume-optimal-policy-in-class}, for any deterministic system $\MDP$ of horizon $\horizon$ and any $T > 0$, the regret for Algorithm~\ref{alg:deterministic} over $T$ rounds of interaction is upper bounded with:
    \begin{align*}
        \regret_T \leq 2 \maxRewards (\horizon + 1) \eluderdim (\policy_\Theta) + 3 \maxRewards \horizon.
    \end{align*}
\end{theorem}

See Section~\ref{subsec:proof-deterministic-main} for the proof of this theorem.

We remark that the $O (\eluderdim (\Theta))$ dependency is generally not improvable for worst-case policy spaces. Suppose, for example, in an extremely large MDP, a finite class of policies without any structure has eluder dimension $|\Theta|$. And in the worst case, the MDP can adversarially make each policy take completely different paths which provides no side-information on other policies, and the agent has to pay at least $\Omega (|\Theta| \maxRewards)$ regret. However, this argument does not characterize any structure in the policy class. In particular, this naive lower bound does not rule out the possibility that a policy class that is easy to learn but has large eluder dimension. In the next subsection, we study the lower bound for an arbitrary policy class with given combinatorial dimensions.

\subsection{A minimax lower bound on the regret}

In this section, we prove minimax lower bounds for the regret depending on the Littlestone dimension. It is important to note that the lower bound holds true for $\emph{any}$ policy class under mild conditions.

We consider the feature vector setup discussed in Section~\ref{sec:combinatorial-dimension}, and assume that $\actionspace = \{0, 1\}$. To formalize the bound, we define the following algorithm class and MDP class:

Given $\horizon \in \mathbb{N}_+, \maxRewards > 0$, feature space $\featSpace$, action space $\actionspace$, and a policy class $\Theta$, to formalize the problem, we denote by $\mathcal{H} (\Theta)$ the class of $\horizon$-epoch deterministic systems whose reward is uniformly bounded by $\maxRewards$ and whose optimal policy lies in $\Theta$. 
Furthermore, we denote by $\mathcal{D}$ the class of deterministic algorithms that act in the MDP environment. An element in $\mathcal{D}$ is a deterministic mapping from the entire observation history to the action space $\actionspace$.

\begin{theorem}\label{thm-deterministic-lower-bound}
    Consider a given time horizon $\horizon \in \mathbb{N}_+$, state-action spaces $(\statespace, \actionspace)$ and feature space $\featSpace$, satisfying $\abss{\actionspace} = 2$ and $\abss{\statespace} \geq 2^{\horizon}$. For any policy space $\Theta$,  we have:
    \begin{align*}
        \inf_{\mathsf{Alg} \in \mathcal{D}} \sup_{\MDP \in \MDPspace (\Theta)} \regret_T (\mathsf{Alg}; \MDP) &\geq \frac{\maxRewards}{4} \left( \littlestonedim (\Theta) \wedge 2^{\horizon} \wedge T - 1 \right).
    \end{align*}
\end{theorem}
See Appendix~\ref{app:lower-bound-proof} for the proof of this theorem. It is worth noticing that Theorem~\ref{thm-deterministic-lower-bound} is valid for \emph{any} policy space generalization model. This means that policy spaces with high Littlestone dimension are fundamentally hard.

Note that Theorem~\ref{thm-deterministic-lower-bound} generalizes the policy learning lower bound in~\cite{du2019good}, where an exponential lower bound is shown for threshold classes. We confirm that this phenomenon is true in general. Since the threshold class has infinite Littlestone dimension, the minimax regret can also be arbitrarily bad.

\section{Proofs}
In this section, we present the proofs of the main theorems. We focus on the proof of Theorem~\ref{thm:stochastic-main}, and also provide an outline for the proof of Theorem~\ref{thm-deterministic-main}. The technical lemmas used in the proof of Theorem~\ref{thm-deterministic-main} and the proof of Theorem~\ref{thm-deterministic-lower-bound} are postponed to Appendices~\ref{app:deterministic-lemmas-proof} and \ref{app:lower-bound-proof}, respectively.

\subsection{Proof of Theorem~\ref{thm:stochastic-main}}\label{app:stochastic-proof}

	 First, we have the following lemma, which guarantees the validity of $Q$ estimator in Algorithm~\ref{alg:stochastic-subroutine} and the optimality gap of the policy induced by the algorithm. This lemma serves as a basic building block, which is used in both the proof of the theorem and the proof of other lemmas.
	 \begin{lemma}\label{lemma:high-confidence-Q-function-bound}
	    Consider Algorithm~\ref{alg:stochastic-subroutine} starting at $(s_{h_0}, a_{h_0}, h_0)$ and the active policy space $\activepolicyspace$. If we have $\thetastar \in \activepolicyspace$, let the event $\Event \mydefn \{\text{line~\ref{algstep-stochastic-q-update} of Algorithm~\ref{alg:stochastic-subroutine} is executed}\}$. We have
	     \begin{align*}
	        \Prob \left( \Event \cap \left\{\abss{\Qhat (s_{h_0}, a_{h_0}, h_0) - \optimalQ (s_{h_0}, a_{h_0}, h_0)} > 2 \maxRewards \sqrt{\frac{\log \delta^{-1}}{N}} \right\} \right) < \delta.
	     \end{align*}
	     Furthermore, if $\Prob \left( \Event \right) > 1 / e$, for any $\theta \in \activepolicyspace$, we have 
	     $\abss{Q^{\policy_\theta} (s_0, a_0, h_0) - \optimalQ (s_0, a_0, h_0)} < \frac{\maxRewards}{N}$.
	 \end{lemma}
	 The proof of this lemma is postponed to Section~\ref{subsubsec:proof-high-confidence-Q}.
	 
	 For the $k$-th round in the outer loop of Algorithm~\ref{alg:policy-elimination-simulator}, we define the following event:
	 \begin{align*}
	     \popEvent_k &\mydefn \{\text{A stack pop operation happens in the $k$-th round}\},\\
	     \pushEvent_k &\mydefn \{\text{A stack push operation happens in the $k$-th round}\},\\
	     \initialEvent_k &\mydefn \{\text{Algorithm~\ref{alg:stochastic-subroutine} starts with $h = 0$}\}.
	 \end{align*}
	 
	 We also define the event $\Event_k \mydefn \pushEvent_k^C$. On the event $\initialEvent_k^C$, Algorithm~\ref{alg:stochastic-subroutine} can be called once or twice in Algorithm~\ref{alg:policy-elimination-simulator} (see line~\ref{algstep:first-call-subroutine} and line~\ref{algstep:second-call-subroutine}), so $\Event_k \cap \initialEvent_k^C$ is the event that line~\ref{algstep-stochastic-q-update} is executed in both calls to Algorithm~\ref{alg:stochastic-subroutine}. On the event $\initialEvent_k$, Algorithm~\ref{alg:stochastic-subroutine} is called once, and $\Event_k \cap \initialEvent_k$ is the event that line~\ref{algstep-stochastic-q-update} is executed in the call to Algorithm~\ref{alg:stochastic-subroutine} starting at $(s_0, a_0, 0)$.
	 
	 For each $k \geq 1$, we consider the event $(\popEvent_k \cup \pushEvent_k \cup \initialEvent_k)^C$. On this event, the stack $\stack$ is non-empty at the beginning of this round, and no push operation is performed in Algorithm~\ref{alg:stochastic-subroutine}. This implies that the stack top is not changed after Algorithm~\ref{alg:stochastic-subroutine} exits, and leads to the stack pop operation in line~\ref{algstep:stochastic-stackpop} of Algorithm~\ref{alg:policy-elimination-simulator}. This cannot happen on the event $(\popEvent_k \cup \pushEvent_k \cup \initialEvent_k)^C$. Therefore, we have:
	 \begin{align}
	     \Prob \left( \bigcup_{k = 1}^T (\popEvent_k \cup \pushEvent_k \cup \initialEvent_k)^C \right) \leq \sum_{k = 1}^T \Prob \left( (\popEvent_k \cup \pushEvent_k \cup \initialEvent_k)^C \right) = 0.\label{eq:proof-of-stoch-exploration-vs-exploitation}
	 \end{align}
	 So with probability one, at least one of the events $(\popEvent_k, \pushEvent_k, \initialEvent_k)$ happens.
	 
	 We use the following lemmas to characterize the behavior of the stack $\stack$.

	  \begin{lemma} \label{lemma-stochastic-stack-vs-eluder}
	       Throughout Algorithm~\ref{alg:policy-elimination-simulator}, at the $k$-th episode, let $(x^{(i)})_{i = 1}^N \subseteq \statespace \times \actionspace^2 \times [ \horizon]$ be the sequence of elements popped out from $\stack$. For any new element $x' \in \statespace \times \actionspace^2 \times [\horizon]$, if $x'$ is pushed into stack $\stack$ at this episode, then $x'$ is independent from $(x^{(i)})_{i = 1}^N$, with respect to $\Theta$ almost surely.
	 \end{lemma}
	 
	 \begin{lemma}\label{lemma-stochastic-depth-of-stack}
    Throughout Algorithm~\ref{alg:policy-elimination-simulator}, $\abss{\stack} \leq \horizon$ always holds almost surely.
    \end{lemma}
    
    Additionally, to guarantee the high-probability validity of the elimination procedure, the following lemma is needed:
    \begin{lemma}\label{lemma-stochastic-optimal-policy-safe}
        With probability $1 - \delta$, throughout $T$ rounds of Algorithm~\ref{alg:policy-elimination-simulator}, we have $\thetastar \in \activepolicyspace_k$ for $k = 1,2, \cdots, T$.
    \end{lemma}
    
    The proof of the three lemmas are postponed to Section~\ref{subsubsec:proof-stochastic-stack-eluder},~\ref{subsubsec:proof-stochastic-depth-stack} and~\ref{subsubsec:proof-stochastic-optimal-safe}, respectively. Based on the three lemmas, we now prove Theorem~\ref{thm:stochastic-main}.
    
    To prove this claim, we recursively construct the item sequence $(y^{(i)})_{i \geq 1} \subseteq \statespace \times \actionspace^2 \times [\horizon ]$ and index sequences $(\ell_i)_{i \geq 1}, (r_i)_{i \geq 1} \subseteq \mathbb{N}$, given the trajectory of stack operations on $\stack$:

\begin{description}
    \item[Step \namedlabel{item:recursive-construction-1}{(1)}] Let $y^{(1)}$ be the first element popped from the stack $\stack$. Let $\ell_1$ be the round at which $y^{(1)}$ is pushed into the stack $\stack$, and $r_1$ be the round at which $y^{(1)}$ is popped from $\stack$.
    \item[Step \namedlabel{item:recursive-construction-2}{($i$)}] For any $i \geq 2$, given $(y^{(j)}, \ell_j, r_j)_{j = 1}^{i - 1}$. Let $y^{(i)}$ be the first element popped from $\stack$ whose push-time is later than $r_{i - 1}$. Let $\ell_i$ be the round at which $y^{(i)}$ is pushed into the stack $\stack$, and $r_i$ be the round at which $y^{(i)}$ is popped from $\stack$.
\end{description}

By the construction, we have:
\begin{align*}
   1 \leq \ell_1 < r_1 \leq \ell_2 < r_2 \leq \cdots \leq \ell_k < r_k < \cdots
\end{align*}

For any $n \geq 1$, define the following function:
\begin{align}
    \numStackEpisodes (n) \mydefn \sum_{i = 1}^n \bm{1}_{\text{A stack operation is performed in the round $i$}}.\label{eq:numStackEpisodes-defn}
\end{align}

Apparently, $\numStackEpisodes$ is a non-decreasing function, which strictly increases at each $\ell_i$ and $r_i$. To control the function $\numStackEpisodes$, we use the following lemma:
\begin{lemma}\label{lemma:stack-sequence}
    Given a stack $\stack$ with operations performed on it through rounds, let $(y^{(i)}, \ell_i , r_i)_{i \geq 1}$ be the items and time points constructed from the operations on $\stack$ according to Step~\ref{item:recursive-construction-1} and Step~\ref{item:recursive-construction-2} for $i \geq 2$. Define the function $\sigma$ Suppose furthermore that $|\stack| \leq L$ is satisfied all the time for some $L > 0$, we have:
    \begin{align*}
        \numStackEpisodes (r_i) = \numStackEpisodes (\ell_i) + 1, \quad \numStackEpisodes (\ell_i) \leq \numStackEpisodes (r_{i - 1}) + 2 L.
    \end{align*}
\end{lemma}

The proof of this lemma is postponed to Section~\ref{subsubsec:proof-stack-seq}.
    
     Define the function $\numStackEpisodes$ according to Eq~\eqref{eq:numStackEpisodes-defn}. Note that we have:
    \begin{align*}
        \numStackEpisodes (T) = \sum_{t = 1}^T \bm{1}_{\popEvent_t} \vee \bm{1}_{\pushEvent_t}.
    \end{align*}

    By Lemma~\ref{lemma:stack-sequence} and Lemma~\ref{lemma-stochastic-depth-of-stack}, for $k \geq 1$, we have:
    \begin{align*}
        \numStackEpisodes (r_k) - \numStackEpisodes (\ell_1) \leq 2 (\horizon + 1) k.
    \end{align*}
    
    Note that since $y^{(1)}$ is the first element popped from $\stack$, and the depth of the stack does not exceed $\horizon$, we have $\numStackEpisodes (\ell_1) \leq \horizon$. Furthermore, according to Lemma~\ref{lemma-stochastic-stack-vs-eluder}, each $y^{(i)}$ constructed in this procedure is independent of $(y^{(j)})_{1 \leq j \leq i}$. Thus, the maximal length $K$ of sequence $(y^{(i)})_{i \geq 1}$ is at most $\eluderdim (\Theta)$. The elements pushed into stack after $r_K$ cannot be popped from the stack (otherwise, a contradiction arises since a new element can be added to the sequence by Step~\ref{item:recursive-construction-2}). Therefore, for any $T \geq 1$, we have:
    \begin{align*}
        \numStackEpisodes (T) \leq 3 \horizon + \numStackEpisodes (r_K) - \numStackEpisodes (\ell_1) \leq 3 \horizon + 2 (\horizon + 1) \eluderdim (\Theta),
    \end{align*}
    almost surely.
    
    On the other hand, by Eq~\eqref{eq:proof-of-stoch-exploration-vs-exploitation}, we know that at least one of $(\popEvent_k, \pushEvent_k, \initialEvent_k)$ happens almost surely. Furthermore, we note since the stack pop operation can only happen in line~\eqref{algstep:stochastic-stackpop} of Algorithm~\ref{alg:policy-elimination-simulator}, we have $\Prob (\popEvent_k \cap \pushEvent_k) = 0$. Consequently, if the stack is initially empty in a round of Algorithm~\ref{alg:policy-elimination-simulator}, the pop operation cannot happen (under $\pushEvent_k$, it cannot happen as discussed above, and under $\pushEvent_k^C$, there is no element to pop from the stack). Therefore, we have:
    \begin{align*}
        \initialEvent_k \setminus (\popEvent_k \cup \pushEvent_k) = \initialEvent_k \setminus \pushEvent_k = \initialEvent_k \cap \Event_k.
    \end{align*}
    
    Combining with the upper bound for $\numStackEpisodes$, we have:
    \begin{align*}
        T - 3 \horizon - 2 (\horizon + 1) \eluderdim (\Theta) \leq T - \numStackEpisodes (T) = \sum_{t = 1}^T \bm{1}_{\initialEvent_t \setminus (\popEvent_t \cup \pushEvent_t)} =  \sum_{t = 1}^T \bm{1}_{\initialEvent_t \cap \Event_t}.
    \end{align*}
    For $T \geq \Tstar = 6 \horizon + 4 (\horizon + 1) \eluderdim (\Theta)$, taking expectations on both sides, we have:
    \begin{align*}
        \frac{1}{T} \sum_{t = 1}^T \Prob \left( \Event_t | \initialEvent_t \right) \geq \frac{1}{T} \sum_{t = 1}^T \Prob \left( \Event_t \cap \initialEvent_t \right) \geq \frac{1}{2}.
    \end{align*}
    Note that the set $\activepolicyspace_k$ is non-increasing throughout the iterations of Algorithm~\ref{alg:policy-elimination-simulator}. Furthermore, for Algorithm~\ref{alg:stochastic-subroutine} starting with the same initial $(s_{h_0}, a_{h_0}, h_0)$ and two different active policy spaces $\Theta_1 \subseteq \Theta_2$, if the state transitions and rewards are coupled together on the same actions, line~\eqref{algstep-stochastic-q-update} is executed with initial active policy space $\Theta_1$ implies that this line is executed with initial active policy space $\Theta_2$. Therefore, the probability $\Prob \left( \Event_t | \initialEvent_t \right)$ is non-decreasing as $t$ increases. So we have:
    \begin{align*}
        \Prob \left( \Event_T | \initialEvent_T \right) \geq \frac{1}{T} \sum_{t = 1}^T \Prob \left( \Event_t | \initialEvent_t \right) \geq \frac{1}{2} > \frac{1}{e}.
    \end{align*}
    The left hand side is the probability that Algorithm~\ref{alg:stochastic-subroutine} starting from $(s_0, a_0, 0)$ with active policy space $\activepolicyspace_T$ executes line~\eqref{algstep-stochastic-q-update}. By Lemma~\ref{lemma-stochastic-optimal-policy-safe}, we have:
    \begin{align*}
        \Prob \left( \thetastar \in \activepolicyspace_T \right) \geq 1 - \delta.
    \end{align*}
    
    On the event $\{\thetastar \in \activepolicyspace_T\}$, invoking Lemma~\ref{lemma:high-confidence-Q-function-bound}, we have:
    \begin{align*}
        \forall \theta \in \activepolicyspace_T, \quad \abss{Q^{\policy_\theta} (s_0, a_0, 0) - \optimalQ (s_0, a_0, 0)} < \frac{\maxRewards}{N} < \varepsilon,
    \end{align*}
    which finishes the proof.

    \subsubsection{Proof of Lemma~\ref{lemma:high-confidence-Q-function-bound}}\label{subsubsec:proof-high-confidence-Q}
    Note that by definition, we have:
    \begin{align*}
        \Event = \bigcap_{i = 1}^N \left\{ \forall h \in [h_0, \horizon], \abss{\policy_\Theta (s_h^{(i)}, h)} = 1  \right\}.
    \end{align*}
    Consider the observation sequence $(s_{h}^{*(i)}, a_h^{* (i)}, \reward_{h}^{* (i)})_{1 \leq i \leq N, h_0 \leq h \leq \horizon}$ generated by taking the optimal policy $\policy_\thetastar$ from the next step of $(s_0, a_0, h_0)$, coupled with the trajectory $(s_{h}^{(i)}, a_h^{(i)}, \reward_{h}^{(i)})_{1 \leq i \leq N, h_0 \leq h \leq \horizon}$ of Algorithm~\ref{alg:stochastic-subroutine} in such a way that for each $i \in [N]$:
    \begin{itemize}
        \item Let $s_{h_0 + 1}^{*(i)} = s_{h_0 + 1}^{*(i)}$ and $R_{h_0}^{(i)} = R_{h_0}^{*(i)}$ almost surely.
        \item If $\abss{\policy_\Theta (s_h^{(i)})} = 1$, let $s_{h + 1}^{*(i)} = s_{h + 1}^{*(i)}$ and $R_{h}^{(i)} = R_{h}^{*(i)}$ almost surely.
        \item If $\abss{\policy_\Theta (s_h^{(i)})} > 1$ at some $h$, couple the path independently afterwards.
    \end{itemize}
    Apparently, on the event $\Event$, we have $(s_{h}^{*(i)}, a_h^{* (i)}, \reward_{h}^{* (i)})_{1 \leq i \leq N, h_0 \leq h \leq \horizon} = (s_{h}^{(i)}, a_h^{(i)}, \reward_{h}^{(i)})_{1 \leq i \leq N, h_0 \leq h \leq \horizon}$.
    By Hoeffding bound, it is easy to see that for any $\delta > 0$, there is:
    \begin{align*}
        \Prob \left( \abss{\frac{1}{N} \sum_{i = 1}^N \sum_{h = h_0}^{\horizon} \reward_h^{* (i)} - \optimalQ (s_{h_0}, a_{h_0}, h_0)} > 2\maxRewards \sqrt{\frac{\log \delta^{-1}}{N}} \right) < \delta.
    \end{align*}
    The first claim then follows by observing the fact that the empirical average of $\reward_h^{* (i)}$ equals $\Qhat (s_{h_0}, a_{h_0}, h_0)$ on the event $\Event$.
    
    Now we prove the second claim. When there is $\Prob (\Event) > 1 / e$, for each $i$, we have:
    \begin{align*}
       \Prob \left(  (s_{h}^{*(i)}, a_h^{* (i)}, \reward_{h}^{* (i)})_{ h_0 \leq h \leq \horizon} = (s_{h}^{*(i)}, a_h^{* (i)}, \reward_{h}^{* (i)})_{ h_0 \leq h \leq \horizon} \right) \geq \Prob\left( \Event \right)^{1 / N} \geq 1 - \frac{1}{N}.
    \end{align*}
    For the $Q$ function, for any $\theta \in \activepolicyspace$, we have that:
    \begin{multline*}
        \abss{Q^{\policy_\theta} (s_0, a_0, h_0) - \optimalQ (s_0, a_0, h_0)} = \abss{\Exs \left(\sum_{h = h_0}^{\horizon} \reward_h^{* (i)} \right) - \Exs \left( \sum_{h = h_0}^{\horizon} \reward_h^{ (i)} \right) }\\
        \leq \Prob \left(  ( \reward_{h}^{* (i)})_{ h_0 \leq h \leq \horizon} \neq (\reward_{h}^{* (i)})_{ h_0 \leq h \leq \horizon} \right) \maxRewards \leq \frac{\maxRewards}{N},
    \end{multline*}
    which proves the second claim.
	 
	\subsubsection{Proof of Lemma~\ref{lemma-stochastic-stack-vs-eluder}}\label{subsubsec:proof-stochastic-stack-eluder}
	
	For the element $x' = (s, a_1, a_2, h) \in \statespace \times \actionspace^2 \times [\horizon]$, suppose $x'$ is pushed into $\stack$ at the $k$-th round of Algorithm~\ref{alg:policy-elimination-simulator}. Note that the stack push operation can only happen at line~\ref{algstep:stochastic-stackpush} of Algorithm~\ref{alg:stochastic-subroutine}, in which case, we have:
	\begin{align}
	    \exists \theta_1, \theta_2 \in \activepolicyspace_k, \quad \policy_{\theta_i} (s, h) = a_i~ \text{for}~ i \in 1,2.\label{eq:proof-stoch-eluder-distinguished-on-push}
	\end{align}

    Consider the stack pop sequence $(x^{(i)})_{i = 1}^N$. Let $k_i < k$ be the episode at which $x^{(i)}$ is popped from $\stack$. Denote $x^{(i)} = (\hat{s}^{(i)}, \hat{a}_1^{(i)}, \hat{a}_2^{(i)}, \hat{h}^{(i)})$, and let $\hat{b}^{(i)} \in \{\hat{a}_1^{(i)}, \hat{a}_2^{(i)}\}$ denote the action being eliminated in the stack pop operation for $x^{(i)}$ (see line~\ref{algstep:stochastic-elimination} of Algorithm~\ref{alg:policy-elimination-simulator}). We have:
    \begin{align*}
        \activepolicyspace_{k} \subseteq \bigcap_{i = 1}^N \left\{\theta \in \Theta: \policy_\theta (\hat{s}^{(i)}, \hat{h}^{(i)} ) \neq \hat{b}^{(i)} \right\}.
    \end{align*}
    
    For $\theta_1, \theta_2 \in \activepolicyspace_k$ defined above and $x^{(i)}$ for $i = 1, 2, \cdots, N$, by above relation, we can conclude that either $\policy_{\theta_1} (\hat{s}^{(i)}, \hat{h}^{(i)}) = \policy_{\theta_2} (\hat{s}^{(i)}, \hat{h}^{(i)})$ or $\{\policy_{\theta_1} (\hat{s}^{(i)}, \hat{h}^{(i)}), \policy_{\theta_2} (\hat{s}^{(i)}, \hat{h}^{(i)})\} \nsubseteq \{\hat{a}_1^{(i)}, \hat{a}_2^{(i)}\}$. Consequently, we have $\theta_1$ and $\theta_2$ are indistinguishable with respect to $x^{(i)}$. By definition, they are also indistinguishable with respect to $(x^{(i)})_{i = 1}^N$. However, by Eq~\eqref{eq:proof-stoch-eluder-distinguished-on-push}, $\theta_1$ and $\theta_2$ are distinguishable with respect to $x'$. Therefore $x'$ is not dependent upon $(x^{(i)})_{i = 1}^N$. The proof is finished.

	\subsubsection{Proof of Lemma~\ref{lemma-stochastic-depth-of-stack}}\label{subsubsec:proof-stochastic-depth-stack}
	
	We first note that in each round of Algorithm~\ref{alg:policy-elimination-simulator}, at most one stack push operation can be performed. This is because Algorithm~\ref{alg:stochastic-subroutine} immediately exits when the stack push operation in line~\ref{algstep:stochastic-stackpush} is invoked, and the second call to Algorithm~\ref{alg:stochastic-subroutine} can happen only when the first call does not change the top of the stack.
	
	 We claim the following fact: at any time, the elements $(s_i, a_{i1}, a_{i2}, h_i)_{i = 1}^{|\stack|}$ from the bottom to the top of the stack has $h_i$ of strictly increasing order. Suppose not, there exists an element $(s, a_1, a_2, h)$, such that when it is pushed to $\stack$, the top element $(s', a_1', a_2', h')$ of the stack satisfies $h' \geq h$. However, $(s, a_1, a_2, h)$ can be pushed into stack only by Algorithm~\ref{alg:optimistic-subroutine}. This is impossible because Algorithm~\ref{alg:optimistic-subroutine} is invoked only starting from stage $h' + 1$.
    
    Therefore, at any time, the elements in $\stack$ has strictly increasing $h$, and we have $|\stack| \leq \horizon$.
	
	\subsubsection{Proof of Lemma~\ref{lemma-stochastic-optimal-policy-safe}}\label{subsubsec:proof-stochastic-optimal-safe}
	Define the events $(\Event_t, \pushEvent_t, \popEvent_t, \initialEvent_t)_{t \geq 1}$ as in the proof of Theorem~\ref{thm:stochastic-main}. Note that for the $t$-th round, the reduction in the active policy space can only happen under event $\popEvent_t \cap \initialEvent_t^C$, which requires the top of the stack to be unchanged under both calls to Algorithm~\ref{alg:stochastic-subroutine} ( line~\ref{algstep:first-call-subroutine} and line~\ref{algstep:second-call-subroutine} in Algorithm~\ref{alg:policy-elimination-simulator}). By the definition of $\Event_t$, it is easy to see that $\popEvent_t \cap \initialEvent_t^C = \Event_t \cap \initialEvent_t^C$.
	
	Let $(s_k, a_{1k}, a_{2k}, h_k)$ be the top element of the stack at the beginning of $k$-th round, and define:
	\begin{align*}
	    \forall i \in \{1, 2\} \quad \Event_k^{(i)} \mydefn \{ \text{line~\ref{algstep-stochastic-q-update} is executed in Algorithm~\ref{alg:stochastic-subroutine} starting with $(s_k, a_{ik}, h_k)$} \}.
	\end{align*}
	We have $\Event_t \cap \initialEvent_t^C = \Event_t^{(1)} \cap \Event_t^{(2)} \cap \initialEvent_t^C$.
	
	If $\policy_\theta (s_{k}, h_{k}) \notin \{a_{1k}, a_{2k}\}$, the optimal policy cannot involve in the elimination. Here we consider the case of $\policy_\theta (s_{k}, h_{k}) \in \{a_{1k}, a_{2k}\}$. Assume $\policy_\theta (s_{k}, h_{k}) = a_{1k}$ without loss of generality.
	By Assumption~\ref{assume-Q-separation}, we have:
	\begin{align*}
	    \optimalQ (s_k, a_{1k}, h_{k}) > \optimalQ (s_k, a_{2k}, h_k) + \Qsep.
	\end{align*}
	
	By Lemma~\ref{lemma:high-confidence-Q-function-bound}, conditionally on $(s_k, a_{1k}, a_{2k}, h_k)$, for any $i \in \{1, 2\}$, we have:
	\begin{align*}
	    \Prob \left( \Event_k^{(i)} \cap \left\{ \abss{\Qhat (s_k, a_{ik}, h_k) - \optimalQ  (s_k, a_{ik}, h_k)} \geq \frac{\Qsep}{2} \right\} \mid s_k, a_{1k}, a_{2k}, h_k \right) \leq 2 \exp \left( - \frac{\Qsep^2 N}{2 \maxRewards^2} \right).
	\end{align*}
	
	Therefore, applying union bound over the events under $\Event_{k}^{(1)}$ and $\Event_k^{(2)}$, on the event $\thetastar \in \activepolicyspace_{k - 1}$, we have:
	\begin{align*}
	    \Prob \left( \Event_k^{(1)} \cap \Event_k^{(2)} \cap \left\{ \Qhat (s_k, a_{1k}, h_k) \leq \Qhat (s_k, a_{2k}, h_k) \right\} \mid s_k, a_{1k}, a_{2k}, h_k \right) \leq 4 \exp \left( - \frac{\Qsep^2 N}{2 \maxRewards^2} \right) < \frac{\delta}{T}.
	\end{align*}
	
	Combining the bounds for all $T$ rounds, we have:
	\begin{align*}
	    \Prob \left( \thetastar \notin \activepolicyspace_{\Tstar} \right) &\leq \sum_{t = 1}^{\Tstar} \Prob \left( \thetastar \notin \activepolicyspace_t, \thetastar \in \activepolicyspace_{t - 1} \right)\\
	    &\leq  \sum_{t = 1}^{\Tstar} \Prob \left( \thetastar \notin \activepolicyspace_t \mid \thetastar \in \activepolicyspace_{t - 1} \right)\\
	    & = \sum_{t = 1}^{\Tstar} \Prob \left( \initialEvent_t^C \cap \Event_t^{(1)} \cap \Event_t^{(2)} \cap \left\{ \Qhat (s_k, a_{1k}, h_k) \leq \Qhat (s_k, a_{2k}, h_k) \right\} \mid \thetastar \in \activepolicyspace_{ t- 1}  \right) \\
	    & < \frac{\delta}{T} \cdot T = \delta,
	\end{align*}
	which finishes the proof.

    \subsubsection{Proof of Lemma~\ref{lemma:stack-sequence}}\label{subsubsec:proof-stack-seq}

    For each $y^{(i)}$, suppose $\numStackEpisodes (r_i) \geq \numStackEpisodes (\ell_i) + 2$, which implies that there are other stack operations performing within the time interval $(\ell_i, r_i)$. Before the episode $r_i$, the element $y^{(i)}$ is not popped from $\stack$ yet. So the stack operations can only involve new elements pushed into the stack after $\ell_i$. However, those elements need to be popped from $\stack$ before $r_i$, which contradicts the fact that $y^{(i)}$ is the first element popped from the stack, who was pushed after $r_{i - 1}$. Therefore, we have $\numStackEpisodes (r_i) = \numStackEpisodes (\ell_i) + 1$.

    On the other hand, after $y^{(i)}$ is popped from the stack, we show that there are at most $2 L$ stack following stack operations before $\ell_{i + 1}$. By Assumption, the depth of $\stack$ never exceeds $L$. Denote by $\tilde{y}^{(i + 1)}$ the first element pushed into the stack after $r_i$, and let $\tilde{\ell}_{i + 1}, \tilde{r}_{i + 1}$ be the episode at which $\tilde{y}^{(i + 1)}$ is pushed into and popped from the stack, respectively. By definition, we have:
    \begin{align*}
        r_i \leq \tilde{\ell}_{i + 1} \leq \ell_{i + 1} < r_{i + 1} \leq \tilde{r}_{i + 1}.
    \end{align*}

    By the minimality of $\tilde{\ell}_{i + 1}$, there is no stack push operation within the interval $[r_i, \tilde{\ell}_{i + 1})$. There are at most $L$ elements that can be popped during this period. Therefore, we have:
    \begin{align*}
        \numStackEpisodes (\tilde{\ell}_{i + 1}) \leq \numStackEpisodes (r_i) + L.
    \end{align*}
    Note that by the FILO property of the stack, within the time period $(\tilde{\ell}_{i + 1}, \tilde{r}_{i + 1})$, the element pushed before $\tilde{\ell}_{i + 1}$ cannot be popped from $\stack$. Note also that $r_{i + 1} \leq \tilde{r}_{i + 1}$. Therefore, within the time interval $(\tilde{\ell}_{ i + 1}, r_{i + 1})$, there are no stack pop operations. Since the depth of the stack cannot exceed $L$, in a time interval at which no pop operations are performed, the number of push operations cannot exceed $L$. Therefore, we obtain:
    \begin{align*}
         \numStackEpisodes (r_{i + 1}) \leq \numStackEpisodes  (\tilde{\ell}_{i + 1}) + L.
    \end{align*}
    Putting them together, we have:
    \begin{align*}
        \numStackEpisodes (\ell_{i + 1}) \leq \numStackEpisodes (r_{i + 1}) \leq \numStackEpisodes  (r_i) + 2 L,
    \end{align*}
    which finishes the proof.
    
\subsection{Proof of Theorem~\ref{thm-deterministic-main}} \label{subsec:proof-deterministic-main}
The proof of Theorem~\ref{thm-deterministic-main} is based on the following three key lemmas:

\begin{lemma}\label{lemma-deterministic-exploration-vs-exploitation}
    For each $k \geq 0$, for the $k$-th episode of Algorithm~\ref{alg:deterministic}, at least one of the following three events happen:
    \begin{itemize}
        \item An element has been pushed into the stack $\stack$.
        \item An element has been popped from the stack $\stack$.
        \item For any $h \in [\horizon]$, $a_h^{(k)} = \policy_{\thetastar}(s_h^{(k)}, h)$.
    \end{itemize}
\end{lemma}

\begin{lemma}\label{lemma-stack-vs-eluder}
    Throughout Algorithm~\ref{alg:deterministic}, at the $k$-th episode, let $(x^{(i)})_{i = 1}^N \subseteq \statespace \times \actionspace^2 \times [ \horizon ]$ be the sequence of elements popped out from $\stack$. For any new element $x' \in \statespace \times \actionspace^2 \times [\horizon ]$, if $x'$ is pushed into stack $\stack$ at this episode, then $x'$ is independent from $(x^{(i)})_{i = 1}^N$, with respect to $\Theta$.
\end{lemma}

\begin{lemma}\label{lemma-depth-of-stack}
    Throughout Algorithm~\ref{alg:deterministic}, $\abss{\stack} \leq \horizon$ always holds.
\end{lemma}

The proof of the lemmas are postponed to Appendices~\ref{subsubsec:proof-exploration-vs-exploitation}, ~\ref{subsubsec:proof-stack-eluder} and~\ref{subsubsec:proof-depth-of-stack}, respectively.

Assuming the three lemmas, now we provide a proof of Theorem~\ref{thm-deterministic-main}. For the $k$-th episode, by Lemma~\ref{lemma-deterministic-exploration-vs-exploitation}, either some stack operation is performed, or the Algorithm~\ref{alg:deterministic} is using the optimal policy $\policy_\thetastar$ throughout all the $\horizon$ epochs. In the latter case, the regret is $0$ for this episode. We note that by assumption, the regret at each episode is uniformly bounded with $\maxRewards$. Therefore, it suffices to show that the first and second scenario in Lemma~\ref{lemma-deterministic-exploration-vs-exploitation} can happen for at most $2 (\horizon + 1) \eluderdim (\policy_\Theta) + 3 \horizon$ episodes of Algorithm~\ref{alg:deterministic}.

Construct the sequence $(y^{(i)})_{i \geq 1} \in \statespace \times \actionspace^2 \times [\horizon]$ and $(\ell_i)_{i \geq 1}, (r_i)_{i \geq 1} \in \mathbb{N}$ recursively according to Step~\ref{item:recursive-construction-1} and Step~\ref{item:recursive-construction-2} as in the proof of Theorem~\ref{thm:stochastic-main}. Note that in the deterministic setting, the time indices are counted by the actual episodes for the learning environment. We can also define the function $\numStackEpisodes$ according to Eq~\eqref{eq:numStackEpisodes-defn}.

Given this lemma, now we get back to the proof of the original theorem. By Lemma~\ref{lemma-depth-of-stack}, we have $|\stack| \leq \horizon$ throughout horizons, which leads to:
    \begin{align*}
        \numStackEpisodes (r_i) = \numStackEpisodes (\ell_i) + 1, \quad \numStackEpisodes (\ell_i) \leq \numStackEpisodes (r_{i - 1}) + 2 \horizon.
    \end{align*}
Let $K$ be the maximal length of the sequence $(y^{(i)})_{i \geq 1}$. The elements pushed into $\stack$ after $r_K$ will never be popped from the stack. By Lemma~\ref{lemma-depth-of-stack}, there are at most $\horizon$ push operations and at most $\horizon$ pop operations after the episode $r_K$. Therefore, there are at most $2 \horizon$ stack operations after $r_K$. On the other hand, the stack operations before $r_1$ are all push operations, which cannot exceed $\horizon$ times. So we have $\numStackEpisodes(\ell_1) \leq \numStackEpisodes(r_1) \leq \horizon$. Putting them together, the number of episodes at which the first and second scenarios in Lemma~\ref{lemma-deterministic-exploration-vs-exploitation} happens is upper bounded by:
\begin{align*}
    \horizon + (\numStackEpisodes (r_K) - \numStackEpisodes(\ell_1)) + 2 \horizon = 3 \horizon + \sum_{i = 1}^K (\numStackEpisodes(r_i) - \numStackEpisodes(\ell_i)) + ( \numStackEpisodes (\ell_i) - \numStackEpisodes(r_{i - 1}) ) \leq 3 \horizon + 2 (\horizon + 1) \eluderdim (\Theta),
\end{align*}
which finishes the proof.

\section{Conclusion and Discussions}
    In this paper, we focus on sample-efficient reinforcement learning in a prohibitively large MDP with a restricted policy space. The notion of eluder dimension is extended to policy spaces, characterizing their intrinsic complexity for learning with exploration. Stack-based exploration algorithms are proposed to learn with policy space generalization. Under a simulator oracle and $\Qsep$-gap in the optimal $Q$ function, we show an $\tilde{O} \left(\horizon \maxRewards \eluderdim (
    \Theta) (\frac{1}{\Delta^2} + \frac{1}{\varepsilon}) \right)$ sample complexity bound for finding an $\varepsilon$-optimal policy. For deterministic systems, the simulator oracle and $Q$-function gap are not needed, and we show a regret upper bound of $O (\horizon \maxRewards \eluderdim (\Theta))$. We also show that the minimax regret is be lower bounded by Littlestone dimension of the policy space. An interesting future direction is to study the possibility of policy space generalization without the simulator and separation condition.
    
\section*{Acknowledgements}
Part of the work was done when Wenlong Mou was a summer intern at Adobe Research. Xi Chen is supported by the NSF Grant via IIS-1845444. The authors would like to thank Georgios Theocharous, Anup Rao, Simon Shaolei Du and Feng Ruan for helpful discussions.

\bibliographystyle{abbrv}
\bibliography{reference}

\appendix

\section*{Appendix}

\subsection*{Organization of the Appendix}
The Appendix is organized as follows: in Section~\ref{app:deterministic-lemmas-proof}, we prove the technical lemmas used in the proof of Theorem~\ref{thm-deterministic-main}; in Section~\ref{app:lower-bound-proof}, we prove Theorem~\ref{thm-deterministic-lower-bound}, the regret lower bound in deterministic systems; finally, in Section~\ref{app:example-proof}, we prove the results related to examples presented in Section~\ref{subsec:examples}.

\section{Proof of technical lemmas in Section~\ref{subsec:proof-deterministic-main}}\label{app:deterministic-lemmas-proof}
In this section, we present proofs of the technical lemmas used in the proof of Theorem~\ref{thm-deterministic-main}, which are postponed from Section~\ref{subsec:proof-deterministic-main} in the main text.

\subsection{Proof of Lemma~\ref{lemma-deterministic-exploration-vs-exploitation}}\label{subsubsec:proof-exploration-vs-exploitation}
    To prove this lemma, we need the following auxiliary lemma:
    \begin{lemma}\label{lemma-Q-function-tournament}
    For any $k \geq 1$, in the $k$-th episode, if an element $(s, a_1, a_2, h) \in \statespace \times \actionspace^2 \times [ \horizon]$ is pushed into $\stack$. At the end of this episode, we have:
    \begin{align*}
        \Qhat (s, a_1, h) = \optimalQ (s, a_1, h),
    \end{align*}
    If an element $(s', a_1', a_2', h')$ is popped from the $\stack$ in the $k$-th episode, at the end of this episode we have:
    \begin{align*}
         \Qhat (s', a_2', h') = \optimalQ (s', a_2', h').
    \end{align*}
    Furthermore, we have $\thetastar \in \activepolicyspace_k$.
    \end{lemma}
    
     The proof of the auxiliary lemmas are postponed to Section~\ref{subsubsec:proof-q-function-tournament}. Assuming Lemma~\ref{lemma-Q-function-tournament}, we now give a proof for Lemma~\ref{lemma-deterministic-exploration-vs-exploitation}.

	Under the condition that no stack push operation is performed during the episode $k$, in the following, we show that the either a stack pop operation is performed, or the policy being executed is optimal from the starting state $s_0$.
	
	We consider the stack at the beginning of episode $k$, which determines the starting state of Algorithm~\ref{alg:optimistic-subroutine}. Suppose the subroutine Algorithm~\ref{alg:optimistic-subroutine} starts from $(s_{h_*}^{(k)}, h_*^{(k)})$. There are two possible cases:
	\paragraph{Case I: $\stack \neq \varnothing$ at the beginning of episode $k$.}
	Let $(s, a_1, a_2, \tilde{h})$ be the top of stack at the beginning of this episode, by definition, we have $\tilde{h} = h_*^{(k)} - 1$, and $\determTran (s, a_2) = s_{h_*^{(k)}}^{(k)}$. Suppose this element is pushed into stack $\stack$ at episode $k_0$. Apparently, we have $k_0 < k$. By Lemma~\ref{lemma-Q-function-tournament}, we have
	\begin{align*}
	    \Qhat (s, a_1, \tilde{h}) = \optimalQ (s, a_1, \tilde{h}),
	\end{align*}
	at the end of $k_0$-th episode. This $Q$-function estimator will not be updated in further episodes, because the dynamics of the system following $(s, a_1, \tilde{h})$ is deterministic and action choices under $\activepolicyspace_k$ are unique, for $k \geq k_0$.
	
	As we have shown, when the stack does not involve push operations in this episode, we also have $\Qhat (s, a_2, \tilde{h}) = \optimalQ (s, a_2, \tilde{h})$. In such case, the elimination step (line~\ref{algstep:deterministic-policy-space-elimination} in Algorithm~\ref{alg:deterministic}) is performed in this round, and a stack pop operation is performed.
	
	\paragraph{Case II: $\stack = \varnothing$ at the beginning of episode $k$.} By the definition of $h_*^{(k)}$, in this case, we have $h_*^{(k)} = 1$. If no element is pushed into $\stack$ in the $k$-th episode, we have:
	\begin{align*}
	     \forall h \in [ \horizon], \quad \abss{\policy_{\activepolicyspace_k} (s_h^{(k)}, h)} = 1.
	\end{align*}
	By Lemma~\ref{lemma-Q-function-tournament}, we have $\thetastar \in \activepolicyspace_k$, which implies that Algorithm~\ref{alg:optimistic-subroutine} performs optimally in this episode.
	
	Therefore, if no new push operations is performed on $\stack$ in this episode, either a pop operation is performed, or the algorithm is following the optimal policy, the proof of this lemma is complete.
	
    \subsection{Proof of Lemma~\ref{lemma-stack-vs-eluder}}\label{subsubsec:proof-stack-eluder}
    For the element $x' = (s, a_1, a_2, h) \in \statespace \times \actionspace^2 \times [ \horizon]$, suppose $x'$ is pushed into $\stack$ at episode $k$. The stack push operation can only happen in two cases: line~\ref{algstep:deterministic-subroutine-push-when-new} in Algorithm~\ref{alg:optimistic-subroutine} and line~\ref{algstep:deterministic-push-in-tournament} in Algorithm~\ref{alg:deterministic}. In both cases, there exists $\theta_1, \theta_2 \in \activepolicyspace_{k}$, such that:
    \begin{align*}
        \policy_{\theta_1} (s, h) = a_1, \quad \policy_{\theta_2} (s, h) = a_2.
    \end{align*}

    Consider the stack pop sequence $(x^{(i)})_{i = 1}^N$. Let $k_i < k$ be the episode at which $x^{(i)}$ is popped from $\stack$. Denote $x^{(i)} = (\hat{s}^{(i)}, \hat{a}_1^{(i)}, \hat{a}_2^{(i)}, \hat{h}^{(i)})$, and let $\hat{b}^{(i)} \in \{\hat{a}_1^{(i)}, \hat{a}_2^{(i)}\}$ denote the action being eliminated in the stack pop operation for $x^{(i)}$ (see line~\ref{algstep:deterministic-policy-space-elimination} of Algorithm~\ref{alg:deterministic}). We have:
    \begin{align*}
        \activepolicyspace_{k} \subseteq \bigcap_{i = 1}^N \left\{\theta \in \Theta: \policy_\theta (\hat{s}^{(i)}, \hat{h}^{(i)} ) \neq \hat{b}^{(i)} \right\}.
    \end{align*}
    
    For $\theta_1, \theta_2 \in \activepolicyspace_k$ defined above and $x^{(i)}$ for $i = 1, 2, \cdots, N$, by above relation, we can conclude that either $\policy_{\theta_1} (\hat{s}^{(i)}, \hat{h}^{(i)}) = \policy_{\theta_2} (\hat{s}^{(i)}, \hat{h}^{(i)})$ or $\{\policy_{\theta_1} (\hat{s}^{(i)}, \hat{h}^{(i)}), \policy_{\theta_2} (\hat{s}^{(i)}, \hat{h}^{(i)})\} \nsubseteq \{\hat{a}_1^{(i)}, \hat{a}_2^{(i)}\}$. Consequently, we have $\theta_1$ and $\theta_2$ are indistinguishable with respect to $x^{(i)}$. By definition, they are also indistinguishable with respect to $(x^{(i)})_{i = 1}^N$. However, by the condition for stack push operation in Algorithm~\ref{alg:deterministic}, $\theta_1$ and $\theta_2$ are distinguishable with respect to $x'$. Therefore $x'$ is not dependent upon $(x^{(i)})_{i = 1}^N$. The proof is finished.
    
\subsection{Proof of Lemma~\ref{lemma-depth-of-stack}} \label{subsubsec:proof-depth-of-stack}
    We claim the following fact: at any time, the elements $(s_i, a_{i1}, a_{i2}, h_i)_{i = 1}^{|\stack|}$ from the bottom to the top of the stack has $h_i$ of strictly increasing order. Suppose not, there exists an element $(s, a_1, a_2, h)$, such that when it is pushed to $\stack$, the top element $(s', a_1', a_2', h')$ of the stack satisfies $h' \geq h$. However, $(s, a_1, a_2, h)$ can be pushed into stack only by Algorithm~\ref{alg:optimistic-subroutine}. This is impossible because Algorithm~\ref{alg:optimistic-subroutine} is invoked only starting from stage $h' + 1$.
    
    Therefore, at any time, the elements in $\stack$ has strictly increasing $h$, and we have $|\stack| \leq \horizon$.

\subsection{Proof of Lemma~\ref{lemma-Q-function-tournament}}\label{subsubsec:proof-q-function-tournament}

	 We prove the result by induction on $k$. Note that $x = (s, a_1, a_2, \tilde{h})$ can be pushed into the stack $\stack$ under two situations: $(i)$, in the line~\ref{algstep:deterministic-subroutine-push-when-new} of Algorithm~\ref{alg:optimistic-subroutine}; and $(ii)$, in the line~\ref{algstep:deterministic-push-in-tournament} of Algorithm~\ref{alg:deterministic}. On the other hand, the element $x' = (s', a_1', a_2', \tilde{h}')$ can be popped from stack $\stack$ in line~\ref{algstep:deterministic-policy-space-elimination} of Algorithm~\ref{alg:deterministic}, which requires the stack to be non-empty and no new element is pushed.
	 
	 For the base case $k = 1$, no stack pop operation can happen in the first episode, and no policy elimination can happen. So we have $\thetastar \in \Theta = \activepolicyspace_1 = \activepolicyspace_2$. For the push operation, only the situation $(i)$ is possible, as the latter case can only happen when there are existing elements in the stack. By the condition for the stack push operation, $h$ is the largest $h$ such that $|\policy_{\activepolicyspace_{k_0}} (s^{k_0}_h, h)| > 1$. Consequently, for $h' > \tilde{h}$, there is $\policy_\theta (s^{k_0}_{h'}, h') = \policy_\thetastar (s^{k_0}_{h'}, h')$. For the updates on $\Qhat$ (line~\ref{algstep:deterministic-q-updates} of Algorithm~\ref{alg:optimistic-subroutine}), each ``max'' operation for $h \in [\tilde{h}, \horizon]$ is actually taken with respect to a singleton, and we have
	 \begin{align*}
	    \forall h \in [\tilde{h}, \horizon], \quad \Qhat (s_h^{(k)}, a_h^{(k)}, h) = \optimalQ (s_h^{(k)}, a_h^{(k)}, h).
	 \end{align*}
	 In particular, note that $a_1 = a_{\tilde{h}}^{(k)}$ by definition, we have $\Qhat (s, a_1, \tilde{h}) =  \optimalQ (s, a_1, \tilde{h})$.
	 
	 Suppose the claim to be true for episodes $1, 2, \cdots, (k - 1)$, we now consider the $k$-th episode, and prove the three claims respectively.
	 
	 \paragraph{$\Qhat$ value for elements pushed into the stack:}
	 	 Both situation $(i)$ and $(ii)$ may happen at the time when $x$ is pushed into the stack. Under situation $(i)$, the arguments for the base case still applies. We now consider the situation $(ii)$: the element $(s, a_1, a_2, \tilde{h})$ is pushed into $\stack$ when running the line~\ref{algstep:deterministic-push-in-tournament} of Algorithm~\ref{alg:deterministic}. Let $(s, b_1', b_2, \tilde{h})$ be the previous element on the top of the stack. There exists $j \in \{1,2\}$, such that $b_j = a_1$. If $j = 1$, the element $(s, b_1', b_2, \tilde{h})$ is pushed into the stack before episode $k$, and the conclusion holds by induction hypothesis. If $j = 2$, we know from Algorithm~\ref{alg:deterministic} that $a_2 = a_{\tilde{h}}^{(k)}$, which is the action taken at $(s, \tilde{h})$ in the $k$-th episode. Since line~\ref{algstep:deterministic-push-in-tournament} is executed in this episode, no stack push operations are performed after horizon $\tilde{h}$. Consequently, we have:
	 \begin{align*}
	    \forall h \in [\tilde{h}, \horizon], \quad \abss{\policy_{\Theta_k} (s_h^{(k)}, h)} = 1.
	 \end{align*}
	 By induction hypothesis, we have $\thetastar \in \activepolicyspace_{k}$, which implies that $\Qhat (s, a_{\tilde{h}}^{(k)}, \tilde{h})  = \optimalQ (s, a_{\tilde{h}}^{(k)}, \tilde{h})$.
	 
	 \paragraph{$\Qhat$ value for elements popped from the stack:}
	 
	 Note by the definition of Algorithm~\ref{alg:deterministic} that in each episode, the Algorithm~\ref{alg:optimistic-subroutine} is called only once. We first assert the following fact: in the $k$-th episode, suppose the subroutine Algorithm~\ref{alg:optimistic-subroutine} starts from $(s_{h_*}^{(k)}, h_*^{(k)})$, and does not involve push operation into the stack, we have:
	\begin{align*}
		\forall h \in [h_*^{(k)}, \horizon], ~\theta \in \activepolicyspace_{k}, \quad a_h^{(k)} = \policy_{\theta} (s_h^{(k)}, h).
	\end{align*}
	This is by the definition of Algorithm~\ref{alg:optimistic-subroutine}: since Algorithm~\ref{alg:optimistic-subroutine} is restricted to choose actions using $\policy_{\activepolicyspace_k}$, suppose the claim is not true, there exists $h > h_*^{(k)},\in \actionspace$ such that the set $\{\policy_\theta (s_h^{(k)}, h) : \theta \in \activepolicyspace_{k} \}$ has cardinality larger than 1, which leads to the push operation for largest such $h$.
	
	By the induction hypothesis, we always have $\thetastar \in \activepolicyspace_k$, and consequently, above expression implies that $a_h^{(k)} = \policy_{\thetastar} (s_h^{(k)}, h)$ for any $h \geq h_*^{(k)}$

	Therefore, if no element is pushed into $\stack$ in $k$-th episode, the current active policy space $\activepolicyspace_{k}$ has no uncertainty on the trajectory after $h_*^{(k)}$. For the updates on $\Qhat$ (line~\ref{algstep:deterministic-q-updates} in Algorithm~\ref{alg:optimistic-subroutine}), each ``max'' operation for $h \in [h_*^{(k)} + 1, \horizon]$ is actually taken with respect to a singleton, and we have
	\begin{align*}
	    \Qhat (s_{h}^{(k)}, \policy_\theta (s_{h}^{(k)}), h) = \optimalQ (s_{h}^{(k)}, \policy_\theta (s_{h}^{(k)}), h), \quad \forall h \in [h_*^{(k)}, \horizon], ~ \theta \in \activepolicyspace_{k}.
	\end{align*}
	In particular, we have:
	\begin{align*}
	    \Qhat (s, a_2, \tilde{h}) = \optimalQ (s, a_2, \tilde{h}).
	\end{align*}
	 
	 \paragraph{The optimal policy $\thetastar \in \activepolicyspace_{k + 1}$:}
    
    If line~\ref{algstep:deterministic-policy-space-elimination} of Algorithm~\ref{alg:deterministic} is not executed in the $k$-th episode, apparently, we have $\activepolicyspace_{k + 1} = \activepolicyspace_k \ni \thetastar$. Now we consider the case where the elimination step is executed, which implies that the stack top is not changed during this episode.
    
    Note that reduction in the policy space can happen only at line~\ref{algstep:deterministic-policy-space-elimination} of Algorithm~\ref{alg:deterministic}. Suppose the $\thetastar$ is eliminated, we have:
	 \begin{align*}
	     \optimalQ (s, \policy_\thetastar (s, h), h) = \Qhat (s, \policy_\thetastar (s, h), h) \leq \Qhat (s, a', h) = \optimalQ (s, a', h),
	 \end{align*}
	 for some $a' \in \policy_{\activepolicyspace_{k}} (s, h) \setminus \{\policy_\thetastar (s, h)\}$, which violates the uniqueness of the optimal policy.
	 
	 Putting them together, the induction proof is finished.

\section{Proof of Theorem~\ref{thm-deterministic-lower-bound}}\label{app:lower-bound-proof}
	 To prove the lower bound, we construct the transition functions, feature vectors, and the reward functions for a given algorithm.
	 
	 Denote $D \mydefn \littlestonedim (\Theta)$. First, without loss of generality, we can assume that $\horizon = \log_2 (D)$. Indeed, if the given time horizon satisfies $\horizon > \log_2 D$, we can construct a deterministic whose first $(\horizon - \lfloor \log_2 D \rfloor)$ epochs do not involve any state transition or rewards, and use a smaller system with $\lfloor \log_2 D \rfloor$ epochs for the rest of the construction. On the other hand, if $\horizon < \log_2 D$, we can simply use only $2^\horizon$ levels in the binary tree and discard the rest.
	 
	 Our construction of the transition structure is based on a binary tree. We first let $\tree$ be a complete binary tree with $\horizon$ layers. For each $h \in [0, \horizon - 1]$, the nodes in the $h$-th level are the set of states reachable at $h$-th epoch of the deterministic system. Without loss of generality, we denote $\statespace = \{0, 1, 2, \cdots, |\statespace| - 1\}$, and let $s_0 = 0$. We construct the transition function:
	 \begin{align*}
	     \determTran (s, a, h) = 2^h \cdot a + s.
	 \end{align*}
	 Clearly, the states reachable at epoch $h$ are $\{0, 1, \cdots, 2^h - 1\}$. We further let the intermediate reward for $h \leq \horizon - 2$ be $0$. In the following, we construct the reward at the horizon $\horizon - 1$ and the feature vectors associated to each $(s, h)$ explicitly based on the trajectory of the algorithm.
	 
	 By the definition of Littlestone dimension, there exists a complete binary tree $(\phi^{(v)})_{v \in \tree}$ of depth $D$ shattered by the policy space $\Theta$.
	 
	 We first assign feature vectors based on the order of states being visited by the algorithm. Note that there are $1 + 2 + \cdots + 2^{\horizon - 1} = D - 1$ possible state-horizon pairs that can be reached. We assign the feature vectors to them according to Algorithm~\ref{alg:adversarial}. Intuitively, when a state is visited for the first time, we record the action in the previous epoch that leads to this state, and take the opposite direction for the path in the tree $\tree$. Using the definition of Littlestone dimension, this path corresponds to an element $\theta \in \Theta$, which is the candidate for the optimal policy in our construction. By this construction, the learning agent is forced to take a sub-optimal action when it visit any $(s, h)$ for the first time, and has to pay for a large amount of regret.
	 \begin{algorithm}[htb]
	     \caption{Adaptive construction of the adversarial MDP}\label{alg:adversarial}
	     \begin{algorithmic}
	     \REQUIRE The trajectory of states visited by a learning agent and a complete binary tree $(\phi^{(v)})_{v \in \tree}$.
	     \ENSURE Feature vectors associated to each $(s, h)$ at its visit time.
	     \STATE Take $v_0 = \tree.\mathrm{root}$ and assign $(\phi^{(v_0)})$ to $(0, 0)$. Initialize $k = 0$.
	     \FOR{each $(s, h)$ visited by the agent}
	     \IF{$(s, h)$ is not visited in the past}
	     \STATE Let $a_k \in \{0, 1\}$ be the action taken to reach $(s, h)$.
	     \STATE Take $v_{k + 1}$ to be child node of $v_k$ at direction $(1 - a_k)$.
	     \STATE Assign the feature vector $\phi^{(v_{k + 1})}$.
	     \STATE Update $k$ with $k + 1$.
	     \ENDIF
	     \ENDFOR
	     \end{algorithmic}
	 \end{algorithm}
	 
	 Now we construct the terminal rewards of the deterministic system. For each terminal state $(s, \horizon - 1)$, if it is the $k$-th terminal state being visited, we let:
	 \begin{align*}
	    \reward (s, \horizon - 1) \mydefn \frac{2 k}{D} \maxRewards. 
	 \end{align*}
	 Apparently, the optimal reward of this deterministic system is the reward at the last terminal state being visited, which is $\maxRewards$. Throughout first $T$ episodes for the agent, the total reward is at most:
	 \begin{align*}
	    \sum_{k = 1}^{T \wedge D / 2} \frac{2k}{D} \maxRewards + \sum_{k \geq D / 2} \maxRewards \leq \begin{cases}
	    \frac{T^2}{D} \maxRewards & T \leq \frac{D}{2},\\
	    \frac{D}{4} \maxRewards + (T - \frac{D}{2}) \maxRewards & T > \frac{D}{2}.
	    \end{cases}
	 \end{align*}
	 The total regret with respect to the optimal policy is therefore lower bounded by $(\frac{D}{4} \wedge \frac{T}{2} ) \maxRewards$. It remains to verify that the optimal policy lies in space $\Theta$.
	 
	 Note that by our construction, the terminal states visited in later episodes always have strictly larger rewards. The optimal policy therefore simply takes the actions opposite to the one chosen for the first time at each $(s, h)$. By the definition of Littlestone dimension, there exists $\theta \in \Theta$, such that:
	 \begin{align*}
	     \forall i \in [D], \quad \policy_\theta (\phi^{(v_{i})}) = 1 - a_i.
	 \end{align*}
	 For $(s, h)$ with $h < \horizon - 1$, when it is visited for the first time, the following state $(s', h + 1)$ is also visited for the first time. Therefore, if $(s, h)$ is associated to the vector $v_i$, the state $(s' h + 1)$ is associated to the vector $v_{i + 1}$, through the action $a_i$, which is the sub-optimal action at $(s, h)$. Since the policy $\theta$ always takes action $1 - a_i$, it is optimal, and the proof for the regret lower bound is finished.

\section{Proofs for the examples} \label{app:example-proof}

In this section, we present proofs of the results about the eluder dimension and Littlestone dimension of specific policy classes, as discussed in Section~\ref{subsec:examples}. We also discuss efficient implementation of the elimination oracle (see Definition~\ref{def:elimination-oracle}) for each example.

\subsection{Threshold function has infinite Littlestone dimensions} \label{subsec:littlestone-infinite}
We prove the claim by direct construction in dimension 1. Any multivariate linear threshold class contains the one-dimensional class as a sub-class, which also has infinite Littlestone dimension.

Let $\theta = [0, 1]$, for arbitrarily large $N$ and any $b \in \{0, 1\}^N$, we can construct the following feature vectors adaptively: choose $\ell_0 = 0$ and $u = 1$. For each $i$, let $\phi^{(i)} = 2 b_i - 1$ and $c_i = \frac{(2 b_i - 1) (\ell_i + u_i)}{2}$. Finally, we update the interval with:
    \begin{align*}
        [\ell_{i + 1}, u_{i + 1}] = \begin{cases} [\ell_i, c_i],& b_i = 0,\\
        [c_i, u_i], & b_i = 1\end{cases}.
    \end{align*}
    For each round $i$, the policies consistent with $(b_j)_{1 \leq j \leq i}$ is the interval $[\ell_i, u_i]$, which has positive length and contains infinite many possible policies. Note that the partition process can be carried out for arbitrarily large $N$. So the Littlestone dimension of the threshold class is larger than any integer, and therefore is infinite.

\subsection{Proof of Proposition~\ref{prop:eluder-example-random-feature}} \label{subsec:eluder-random-feature}
Let $(\phi_i)_{i = 1}^{D} \sim \mathrm{i.i.d.} \mathcal{N} (0, I_d)$. For any $\theta_1, \theta_2 \in \Theta$, since $\vecnorm{\theta_1 - \theta_2}{2} \geq \varepsilon$ and $\theta_1, \theta_2 \in \sphere^{d - 1}$, there is:
\begin{align*}
    \Prob \left( \policy_{\theta_1} (\phi_i) \neq \policy_{\theta_2} (\phi_i) \right) \geq \frac{2 \arcsin \frac{\varepsilon}{2}}{2 \pi} \geq \frac{\varepsilon}{2 \pi}.
\end{align*}
By independence, we have:
\begin{align*}
    \Prob \left( \forall i \in [D], \policy_{\theta_1} (\phi_i) = \policy_{\theta_2} (\phi_i) \right) = \Prob \left( \policy_{\theta_1} (\phi_1) = \policy_{\theta_2} (\phi_1) \right)^D \leq \left(1 - \frac{\varepsilon}{2 \pi} \right)^D \leq \exp \left( - \frac{\varepsilon D}{2 \pi} \right).
\end{align*}
Taking union bound over $\binom{|\Theta|}{2}$ possible pairs, we obtain:
\begin{align*}
    \Prob \left( \exists \theta_1 \neq \theta_2 \in \Theta, \policy_{\theta_1} (\phi_i) = \policy_{\theta_2} (\phi_2),~\forall i \in [D]\right) \leq |\Theta|^2 \exp \left( - \frac{\varepsilon D}{2 \pi} \right).
\end{align*}
Taking $D = \frac{4 \pi}{ \varepsilon} \log \frac{|\Theta|}{\delta}$, the probability of above event is at most $\delta$. On this event, for any $\phi' \in \sphere^{d - 1}$, we have $\phi'$ is independent of $(\phi_i)_{i = 1}^D$ with respect to $\Theta$, which proves the claim.

\subsection{Proof of Proposition~\ref{prop:eluder-gf2}} \label{subsec:proof-gf2}
We first prove an upper bound on the eluder dimension, and then prove a lower bound on the Littlestone dimension.

Given $\phi_1, \phi_2, \cdots, \phi_{k} \in \gftwo^D$, if $\phi_k$ independent of $(\phi_j)_{1 \leq k - 1}$ with respect to $\Theta$. By definition, there exists $\theta_1, \theta_2 \in \gftwo^D$, such that $\inprod{\phi_j}{\theta_1} = \inprod{\phi_j}{\theta_2}$ for any $j \in \{1, 2, \cdots, k - 1\}$ but $\inprod{\theta_1}{\phi_k} \neq \inprod{\theta_2}{\phi_k}$, which implies that $\phi_k$ cannot be linearly represented by $(\phi_j)_{j = 1}^{k - 1}$ in $\gftwo$. For a sequence of $m$ vectors $\phi_1, \phi_2, \cdots, \phi_m$, since $\phi_i$ cannot be linearly represented by $\phi_1, \phi_2, \cdots, \phi_{i - 1}$ for each $i$, the set of vectors are linearly independent. In a $D$-dimensional vector space, the length of this sequence cannot exceed $D$. So we have $\eluderdim (\Theta) \leq D$.

On the other hand, let $\phi_i = e_i$, the vector with $1$ at $i$-th entry and $0$-s elsewhere. For a given binary sequence $b_1, b_2, \cdots, b_D$, let $\theta = (b_1, b_2, \cdots, b_D) \in \gftwo^D$, we have:
\begin{align*}
    \forall i \in [D], \quad \policy_\theta (\phi_i) = \inprod{\theta}{\phi_i} = b_i.
\end{align*}
Construct a $\gftwo^D$-valued binary tree with all the nodes at $i$-th level being $\phi_i$. By the definition of Littlestone dimension, we have $\littlestonedim (\Theta) \geq D$.

Putting them together, and noting that $\littlestonedim (\Theta) \leq \eluderdim (\Theta)$, we finish the proof of this proposition.

\subsection{Proof of Proposition~\ref{prop:eluder-fourier}} \label{subsec:proof-fourier-boolean}
The proof is similar to that of Proposition~\ref{prop:eluder-gf2}, but the arithmetics are carried out under real numbers, instead of $\gftwo$.

Given $\phi_1, \phi_2, \cdots, \phi_{k} \in \{-1, 1\}^D$, for each $\phi_j$, we define the vector $v_j \mydefn [\chi_S (\phi_j)]_{S \subseteq \mathcal{A}} \in \real^{|\mathcal{A}|}$. If $\phi_k$ independent of $(\phi_j)_{1 \leq k - 1}$ with respect to $\mathcal{F} (\mathcal{A})$. By definition, there exists $f_1, f_2 \in \mathcal{F} (\mathcal{A})$, such that $f_1 (\phi_j) = f_2 (\phi_j)$ for any $j \in \{1, 2, \cdots, k - 1\}$ but $f_1 (\phi_k) \neq f_2 (\phi_k)$. Note that by the Fourier expansion, $f_i (x) = \sum_{S \in \mathcal{A}} \hat{f}_i (S) \chi_S (x) = \inprod{[\hat{f}_i (S)]_{S \in \mathcal{A}}}{v_j}$. This implies that $v_k$ cannot be linearly represented by $(v_1, v_2, \cdots, v_{k - 1})$ with real coefficients. For a sequence of $m$ vectors $v_1, v_2, \cdots, v_m$, since $v_i$ cannot be linearly represented by $v_1, v_2, \cdots, v_{i - 1}$ for each $i$, the set of vectors are linearly independent. In a $|\mathcal{A}|$-dimensional vector space, the length of this sequence cannot exceed $|\mathcal{A}|$. So we have $\eluderdim (\Theta) \leq |\mathcal{A}|$.

\subsection{Efficient implementation of the elimination oracle}\label{app:efficient-oracle-examples}
In this section, we present computationally efficient algorithms for the elimination oracle (see Definition~\ref{def:elimination-oracle}) for the examples discussed in Section~\ref{subsec:examples}.

\paragraph{Linear threshold functions:} Consider the linear threshold policy class discussed in Example~\ref{example:linear-worst-case}. For a set of feature vectors $Z = (\phi_1, \phi_2, \cdots, \phi_M) \subseteq \real^d$, signs $(b_1, b_2, \cdots, b_M) \subseteq \{\pm 1\}$, as well as a new feature vector $\phi'$, the elimination oracle requires finding the possible signs of $\theta^\top \phi'$, such that $\forall k, ~\mathrm{sgn} (\theta^\top \phi_k) \neq b_k$. This is equivalent to the solvability of the following linear programs:
\begin{align*}
    \begin{cases}
    \theta^\top \phi' > 0,\\
    b_1 (\theta^\top \phi_1) < 0,\\
    \cdots\\
    b_M (\theta^\top \phi_M) < 0.
    \end{cases}
    \quad \mbox{and} \quad
     \begin{cases}
    \theta^\top \phi' < 0,\\
    b_1 (\theta^\top \phi_1) < 0,\\
    \cdots\\
    b_M (\theta^\top \phi_M) < 0.
    \end{cases}
\end{align*}
It is known that they can be solved in $\mathrm{poly} (d, M)$ time.

\paragraph{Linear functions in $\gftwo^d$:} Consider the linear function class in $\gftwo^d$ described in Example~\ref{example:gf2}. For a sequence of feature vectors $Z = (\phi_1, \phi_2, \cdots, \phi_M) \subseteq \real^D$, bits $(b_1, b_2, \cdots, b_M) \subseteq \gftwo$, as well as a new feature vector $\phi'$, the elimination oracle requires finding the possible value of $\inprod{\phi'}{\theta}$ such that $\forall k, ~\inprod{\theta}{b_k} \neq b_k$. This is equivalent to the solvability of the following linear systems in $\gftwo$:
\begin{align*}
    \begin{cases}
    \inprod{\theta}{\phi'} = 1,\\
    \inprod{\theta}{ \phi_1} = 1 - b_1,\\
    \cdots\\
    \inprod{\theta}{ \phi_M} = 1 - b_M.
    \end{cases}
    \quad \mbox{and} \quad
    \begin{cases}
    \inprod{\theta}{\phi'} = 0,\\
    \inprod{\theta}{ \phi_1} = 1 - b_1,\\
    \cdots\\
    \inprod{\theta}{ \phi_M} = 1 - b_M.
    \end{cases}
\end{align*}
Applying Gaussian elimination algorithm in $\gftwo$ solves the equation systems within $\mathrm{poly} (D, M)$ time.

For the Fourier-concentrated functions discussed in Example~\ref{example:discrete-fourier}, we can formulate a pair of $|\mathcal{A}|$-dimensional real-valued linear systems in the same way, and they can also be solved by Gaussian elimination within $\mathrm{poly} (|\mathcal{A}|, M)$ time.
\end{document}